\newif\ifcameraready
\camerareadytrue   %% <-- EDIT THIS LINE TO SWITCH MODES

\ifcameraready
  \documentclass[acmtog]{acmart}
\else
  \documentclass[acmtog, anonymous, review]{acmart}
\fi

\usepackage{xcolor}
\usepackage{amsmath}
\usepackage{mathrsfs}
\makeatletter
\renewcommand\@dbflt[1]{\@ifnextchar[{\@xdblfloat{#1}}{\@xdblfloat{#1}[tbp]}}
\makeatother

\usepackage{graphicx}
\usepackage{caption}
\usepackage{array}
\usepackage{adjustbox}

\AtBeginDocument{%
  \providecommand\BibTeX{{%
    \normalfont B\kern-0.5em{\scshape i\kern-0.25em b}\kern-0.8em\TeX}}}

\setcopyright{acmlicensed}
\copyrightyear{2026}
\acmYear{2026}
\acmConference[SIGGRAPH '26 Conference Proceedings]{ACM SIGGRAPH 2026 Conference Proceedings}{July 19--23,
  2026}{Los Angeles, CA, USA}
\ifcameraready\else
  \acmSubmissionID{1156}
\fi

\acmJournal{TOG}
\acmVolume{0}
\acmNumber{0}
\acmArticle{}
\acmYear{2026}
\acmMonth{5}
\setcopyright{rightsretained} 
\definecolor{Gray}{rgb}{0.5,0.5,0.5}
\definecolor{darkblue}{rgb}{0,0,0.7}
\definecolor{darkgreen}{rgb}{0.1,0.5,0.1}
\definecolor{orange}{rgb}{1,.5,0} % something readable but different from todo
\definecolor{red}{rgb}{1,0,0} % something readable but different from todo

\newcommand{\relatedpar}[1]{\vspace{0.3mm}\noindent\textbf{#1.}}

\newcommand{\rev}[1]{#1}

\newif\ifdraft
\drafttrue
\ifdraft

\newcommand{\bl}[1]{{\color{purple}#1}}

\else
\newcommand{\bl}[1]{{\color{black}#1}}
\fi

\newcommand{\ignorethis}[1]{}

\newcommand{\sectnum    } [1] {\ref{#1}}
\newcommand{\tblnum     } [1] {\ref{#1}}
\newcommand{\fignum     } [1] {\ref{#1}}

\newcommand{\sect       } [1] {Section~\sectnum{#1}}
\newcommand{\sects      } [1] {Sections~\sectnum{#1}}
\newcommand{\tbl        } [1] {Table~\tblnum{#1}}

\newcommand{\fig        } [1] {Figure~\fignum{#1}}

\newcommand{\etal       }     {{et~al.}}

\begin{document}

\title{Lucky High Dynamic Range Smartphone Imaging}

%% ============================================================
%%  AUTHORS.  In camera-ready mode the full author block is emitted.
%%  In submission mode acmart itself hides authors via the `anonymous'
%%  option, so we leave this block guarded by \ifcameraready.
%%  ORCID and email placeholders should be replaced with real values
%%  before the final camera-ready submission.
%% ============================================================
\ifcameraready
\author{Baiang Li}
\authornote{Equal contribution.}
\email{baiang.li@princeton.edu}
\affiliation{%
  \institution{Princeton University}
  \city{Princeton}
  \state{NJ}
  \country{USA}
}

\author{Ruyu Yan}
\authornotemark[1]
\email{ruyu.yan@princeton.edu}
\affiliation{%
  \institution{Princeton University}
  \city{Princeton}
  \state{NJ}
  \country{USA}
}

\author{Ethan Tseng}
\email{eftseng@princeton.edu}
\affiliation{%
  \institution{Princeton University}
  \city{Princeton}
  \state{NJ}
  \country{USA}
}

\author{Zhoutong Zhang}
\email{zhoutongz@adobe.com}
\affiliation{%
  \institution{Adobe}
  \country{USA}
}

\author{Adam Finkelstein}
\email{af@princeton.edu}
\affiliation{%
  \institution{Princeton University}
  \city{Princeton}
  \state{NJ}
  \country{USA}
}

\author{Jiawen Chen}
\email{jiawen@waabi.ai}
\affiliation{%
  \institution{Adobe}
  \country{USA}
}

\author{Felix Heide}
%\authornote{Corresponding author.}
\email{fheide@princeton.edu}
\affiliation{%
  \institution{Princeton University}
  \city{Princeton}
  \state{NJ}
  \country{USA}
}
\fi

%%
%% Short author list for page headers.
\ifcameraready
  \renewcommand{\shortauthors}{Li, Yan, Tseng, Zhang, Finkelstein, Chen, and Heide}
\else
  \renewcommand{\shortauthors}{Anon. Submission ID: 1156}
\fi

\begin{abstract}
While the human eye can perceive an impressive twenty stops of dynamic range, smartphone camera sensors remain limited to about twelve stops despite decades of research. A variety of high dynamic range (HDR) image capture and processing techniques have been proposed, and, in practice, they can extend the dynamic range by 3-5 stops for handheld photography. This paper proposes an approach that robustly captures dynamic range using a handheld smartphone camera and lightweight networks suitable for running on mobile devices.
Our method operates indirectly on linear raw pixels in bracketed exposures. 
Every pixel in the final HDR image is a convex combination of input pixels in the neighborhood, adjusted for exposure, and thus avoids hallucination artifacts typical of recent deep image synthesis networks.
We validate our system on both synthetic imagery and unseen real bracketed images -- we confirm zero-shot generalization of the method to smartphone camera captures.
Our iterative inference architecture is capable of processing an arbitrary number of bracketed input photos, and we show examples from capture stacks containing 3--9 images.
Our training process relies only on synthetic captures yet generalizes to unseen real photos from several cameras. Moreover, we show that this training scheme improves other SOTA methods over their pretrained counterparts. % and proposed training approaches.
\ifcameraready
\textbf{Project page:} \textcolor{blue}{\url{https://lucky-hdr.github.io/}}.\par\medskip
\fi
\end{abstract}
%  - alignment for bracketed captures
%  - merging non local patches
%  - depart from denoising and focus on pixel selection
%  - operate indirectly on pixels to avoid hallucination and CNN artifacts
%  - validate on simulated and real captures




\begin{teaserfigure}
\centering
    %\vspace*{-0.2\baselineskip}
    \begin{minipage}[t]{1.0\linewidth}
        \centering
        \captionsetup{font={tiny}}
        \includegraphics[height=4cm,width=18cm]{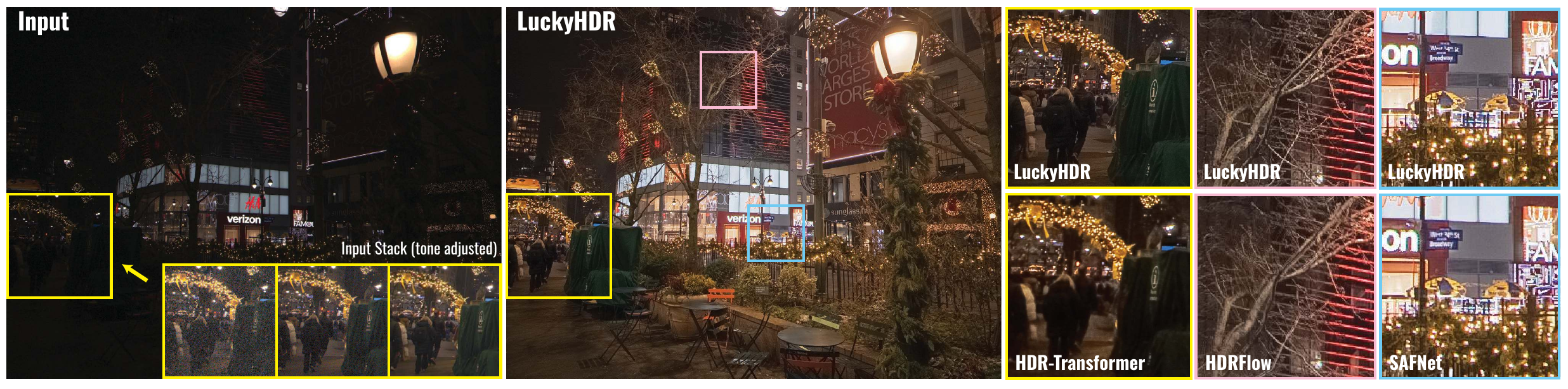}
    \end{minipage}
    \vspace*{-1.5\baselineskip}
\caption{\textbf{LuckyHDR Imaging.} Left: captured handheld bracketed burst of 6 exposures (in [-3.0, +2.0] EV, subset inset). Middle: our method aligns and merges the input frames to produce a clean HDR image. Right: zoomed regions compare our method (above) to recent baseline methods 
(HDR-Transformer~\cite{liu2022ghost},
HDRFlow~\cite{xu2024hdrflow},
SAFNet~\cite{kong2025safnet}, below),
all of which operate on a fixed [-2.0, +2.0] EV range. LuckyHDR generalizes to an arbitrary-length stack covering a wider exposure range -- achieving higher fidelity while avoiding hallucinations (by design) and misalignment artifacts.}
  \label{fig:teaser}
\end{teaserfigure}

\maketitle

\vspace{-3mm}
\section{Introduction}

While smartphones have become ubiquitous personal cameras, they often fail to capture the world as we perceive it, most notably in high dynamic range (HDR) scenery.
The human eye can perceive about twenty stops of dynamic range~\cite{darmont2013high}, whereas modern smartphones are limited to about twelve stops because a small sensor form factor implies both high noise and a low saturation ceiling.
To capture HDR imagery on a smartphone, this paper takes inspiration from an approach in astrophotography called \emph{lucky imaging}, wherein atmospheric artifacts can be 
avoided by combining ``lucky'' pixels from a burst of short exposures
\cite{fried1978probability}.
Over the years, researchers have developed numerous HDR photography methods. Traditional \emph{exposure bracketing} captures multiple images with varying exposure and/or gain (ISO), but typically requires a tripod and static scene. While computational methods for aligning and merging handheld image sequences have been proposed (e.g. \cite{hasinoff2016burst}), they struggle at both extremes -- shadows and highlights.
Alternative approaches include spatially varying exposures and optical masks to capture several exposure stops in a single frame, trading spatial resolution for dynamic range. More recently, HDR image inpainting techniques use deep learning to recover missing highlights and shadows, but these can only reconstruct a few stops and suffer from hallucination artifacts.

Our method combines bracketed exposures from handheld mobile photography to achieve high dynamic range while avoiding hallucination.
%on par with human vision. \fig{fig:difference} illustrates the proposed method compared to existing approaches. 
Rather than directly predicting pixel values, we train deep learning models that operate indirectly on the input by aligning the bracketed ``stack'' of raw pixels and interpolating between their values. Even after alignment, many observed pixels cannot be naively merged due to noise in dark regions, saturation in bright regions, motion blur in long exposures, or alignment failures from (dis)occlusions. Our key insight is that we only need one or a few ``lucky'' captured pixels in each region to estimate the color at that location. We show that our model, trained entirely via simulated data, can identify and interpolate these lucky pixels.

While our task involves optical flow registration, we focus on scenarios with small scene motion and modest hand shake. These scenarios are typical in multi-frame computational photography tasks like scene reconstruction, multi-exposure fusion, and radiance field acquisition. 
We introduce a new synthetic raw dataset that faithfully simulates hand shake and motion blur to train a joint alignment and merging model that generalizes to real raw data, avoiding time-consuming and error-prone real-world data collection.
In addition to training our models, we validate that this dataset improves the zero-shot quality of the best existing baseline methods over their pretrained counterparts -- all tested methods are evaluated on unseen captures from  smartphone cameras and synthetic data.
Our models are capable of fusing bracketed image stacks of 3--6 photos captured by smartphone to produce HDR images of favorable quality to SOTA methods while being smaller and faster, making them suitable for mobile deployment.
We also validate the method with with an extreme synthetic exposure stack of 9 images spanning more than 20 stops -- on par with the limits of human vision.

This paper includes the following contributions:
\begin{itemize}
\item A system capable of capturing SOTA quality HDR scenery using the camera and processing available on a smartphone.
\item A flexible iterative fusion architecture that can handle an arbitrary number of input images of varying exposure.
\item Models trained on synthetic raw images that generalize to real photos captured using a smartphone camera.
\item A pixel-combination approach that avoids hallucination artifacts common in direct pixel synthesis methods.
\end{itemize}

\noindent Source code, models, and data are available in supplemental materials and will be made public upon publication.
  
% Today's HDR in smartphones
%%% Bracketed Image Capture
%%% HDR as denoising

% Today's HDR in deep learning / research
%%% HDR hallucination (Thomas' work)
%%% Optical masks

% In this work
%%% Indirect operation on pixels
%%% Avoid hallucination
%%% Convert denoising into pixel picking through long exposures
%%% Investigate bracketed alignment to handle small amounts of parallax and occlusion
%%% Investigated nonlocal patches merging
%%% Develop a smartphone app for bracketed capture
%%% Trained only in simulation
%%% Evaluate on simulated and real world brackets

% Limitations
%%% Relies on a static scene
%%% Somewhat long-ish capture time
%%% Still need to investigate tone mapping

%\input{figs_and_tabs/figure_difference}

\vspace*{-3mm}
\section{Related Work}

We next summarize prior work in high dynamic range photography.
%\adam{Felix mentioned AFUNet on slack.}
%\adam{Are there new papers that came out since our SA25 submission? Were there papers discussed in rebuttal that we should add here?}
%--> Baiang to add recent gen methods

\relatedpar{Multi-frame HDR Photography} 
Early HDR techniques use exposure bracketing with tripod-mounted cameras in static scenes~\cite{mann1995undigital,debevec1997hdr}. Later work extends these methods to handle handheld camera motion~\cite{ward2003fast}, scene motion~\cite{gallo2009artifact}, or both~\cite{zimmer2011freehand,sen2012patch}, while parallel research investigates optimal frame selection~\cite{granados2010optimal,hasinoff2010noise,gallo2012metering} and perceptual quality metrics~\cite{mantiuk2011hdrvdp2,tursan2016objective}.
Nevertheless, despite decades of research, most methods struggle with practical smartphone photography. HDR+~\cite{hasinoff2016burst,liba2019verylow} improves reliability but originally does not perform bracketing, limiting its dynamic range. Recent extensions incorporate some bracketing~\cite{ernst2021bracketing} and deep learning~\cite{niu2021hdr,yan2022dual,lecouat2022hdr}, but still struggle with extreme highlights and shadows.

\relatedpar{Single-frame HDR Prediction}
To avoid multi-frame alignment challenges, single-frame methods reconstruct HDR content using priors based on sparse features~\cite{serrano2016sparse,fotiadou2020snapshot} or deep learning~\cite{eilertsen2017single,kalantari2017deep,liu2020reverse,santos2020perceptual,eilertsen2021cheat,zou2023rawhdr}. These methods typically extend dynamic range by 1-2 stops.

\relatedpar{HDR Sensors}
Robotic applications like self-driving vehicles require real-time HDR sensing. Existing approaches use temporal or spatial multiplexing of different exposures and gains \cite{Mann1994BeingW,Reinhard2010HighDR}. Temporal multiplexing introduces motion artifacts~\cite{Grossberg2003HighDR,mertens2009exposure,hasinoff2010noise}, while split-pixel designs~\cite{Deegan2014,Takayanagi2018,Tanaka2018} require large sensors that are not suitable for smartphones. Recent dual conversion gain sensors~\cite{park20190,venezia20181} achieve up to 
%90dB dynamic range but trade latency for range.
about 15 stops at the expense of sensor cost and pixel size, making these sensors approaches unsuitable for mobile photography.

\relatedpar{Alternative Single-shot HDR Approaches}
Novel sensor designs include spatially varying exposures~\cite{nayar2000spatially}, modulo cameras~\cite{zhao2015unbounded,so2021mantissa}, event cameras~\cite{han2020neuromorphic,messikommer2022event}, and optical encodings~\cite{sun2020rank1,metzler2020deep}. While promising, these alternative sensors have not yet been commercialized in smartphones.

\relatedpar{Learning-based HDR Reconstruction and Generative HDR} The most successful recent methods for exposure-bracketed HDR reconstruction investigate learning-based approaches using attention and optical flow, including AHDRNet~\cite{yan2019attention}, SAFNet~\cite{kong2025safnet}, HDRFlow~\cite{xu2024hdrflow}, HDR-Transformer~\cite {liu2022ghost}, and AFUNet~\cite{li2025afunet}. These methods methods typically predict HDR pixels directly as a network output or rely on learned fusion to suppress misalignment artifacts. We compare against these methods in \sect{sec:evaluation}, and, as such, find that they often exhibit hallucinated texture or ghosting for challenging dynamic scenes or extreme exposure differences.
Recent diffusion-based approaches, e.g., for event-guided HDR reconstruction~\cite{yang2025event} or HDR video reconstruction~\cite{guan2024diffusion} have also been shown to achieve visually convincing results by leveraging from generative priors. These methods are by design prone to hallucinations and their computational cost prohibits mobile on-device HDR reconstruction. In contrast, LuckyHDR follows a measurement-driven design: we predict only local shifts and merging weights, and reconstruct the output via  warping and compositing of observed pixels.

\nocite{Phillips:2018:CIQ:3208704,yahiaouiimpact,schulz2007using,battiato2010image,vuong2008new,park2009method,su2015fast,su2016model,shim2018gradient,ding2020adaptive}
\nocite{li2025afunet,chen2025robust} %These methods comes out in iccv 2025. 

\begin{figure*}[t]
\vspace{-4mm}
	\centering
	\includegraphics[width=0.99\textwidth]{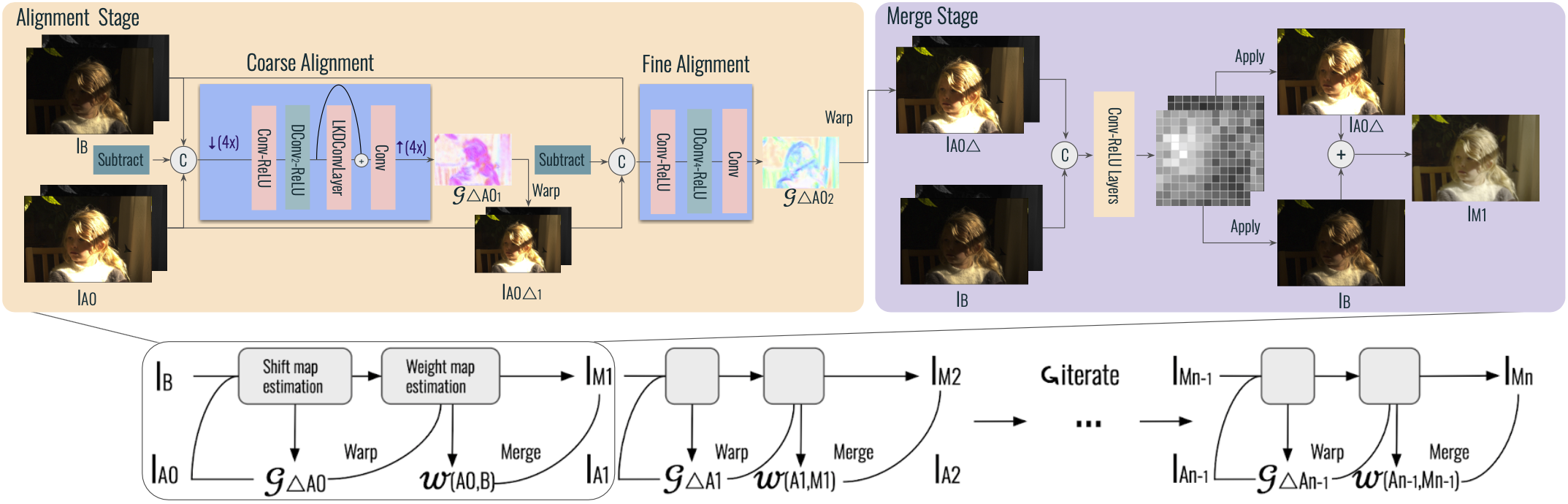}\vspace{-2mm}
 	\caption{\textbf{LuckyHDR Overview.} 
    LuckyHDR merges images sequences of arbitrary length by iterating an align-and-merge pipeline.
    At each iteration, the Shift Stage (yellow) calculates a map that locally aligns the new frame $I_{Ai}$ onto the base frame $I_B$ or intermediate result $I_{Mi}$.
    The Merge Stage (blue) calculates a weight map for blending the aligned frame $I_{Ai}$ and $I_B$ (or $I_{Mi}$), to produce the next merged result $I_{Mi+1}$. 
    The shift module uses shared weights across this chain, as does the merge module.
    This architecture can handle an arbitrary number of input frames 
    and does not require retraining for each sequence length. Here, DConv$_2$-ReLU (second layer from left) and DConv$_4$-ReLU represent dilated convolution with dilation 2 and 4, respectively, and followed by a ReLU activation layer. LKDConvLayer represents a convolutional layer with Dilation=3 and kernel size=7, followed by a Conv-1 layer with kernel size = 1.
    }
 \label{fig:luckyhdr-overview}
 %\vspace{-3mm}
\end{figure*}

\vspace*{-3mm}
\section{Lucky Capture, Align and Merge}
\label{sec:method}

We introduce our proposed LuckyHDR image reconstruction method, which is illustrated in \fig{fig:luckyhdr-overview}. We capture a sequence of bracketed frames 
%(metered by an autoexposure algorithm) 
and recover and HDR image iteratively \emph{as frames stream in}. This section first formalizes the image formation model (\sect{sec:formation}) and then describes our automatic exposure and capture procedure (\sect{sec:capture}). The proposed method iteratively aligns and merges from the shortest exposure (the \emph{base frame}) to the longest, gradually blending in reduced-noise texture in mid-tones and finally the shadows. To do so, the method alternates between: (\sect{sec:align}) an alignment stage that computes a local shift map compensating for camera and scene motion; and (\sect{sec:merge}) a merge stage  that predicts a weight map used to merge in image content not adequately captured in earlier iterations.
Both steps employ lightweight neural networks, whose weights do not change across iterations. We emphasize that our method does not directly predict pixel values. 

The short-to-long iterative align and merge process shown in \fig{fig:luckyhdr-overview} supports an elegant user experience: the user begins with a usable (but noisy) viewfinder image that corresponds to the moment in time. The system gradually improves the result until the user decides to stop.

\subsection{Image Formation Model}
\label{sec:formation}

Next, we formalize the HDR reconstruction problem. A single camera sensor pixel records light following
\begin{equation}
\label{eq:img-formation}
    V = \text{round}(\min(A (p(I \Delta t) + \eta(\sigma)), T)),
\end{equation}
where $V$ is the recorded digital value~\cite{tseng2021optics}.
The photon flux $I$ is determined by scene radiance and camera optics; and given exposure time $\Delta t$, the analog signal follows a Poisson distribution $p$. The variance of this distribution is typically referred to as shot noise. The other major noise contribution is read noise, which is typically modeled as a Gaussian $\eta(\sigma) \sim \mathcal{N}(0,\sigma_r)$, although more accurate noise models can also be used \cite{wei2020physics}. The analog signal is amplified by a gain $A$ (camera ISO) before it is clamped to well capacity $T$ and quantized to a digital number.
For smartphone cameras, dynamic range is limited due to the small sensor size.
A low well capacity $T$ limits the maximum intensity achievable.
Likewise, analog gain $A$ needs to be large to sense the signal, amplifying an already high read noise.

Bracketed photography ameliorates noise and increases dynamic range by fusing a series of captures $V_1(A_1, \Delta t_1), \dots, V_n(A_n, \Delta t_n)$ with varying exposure and gain, combined as a weighted average
\begin{equation}
    H = \sum_{i=1}^n w_i V_i(A_i, \Delta t_i).
\end{equation}
In any particular frame, some pixels will generally be underexposed and therefore noisy, while other pixels will be overexposed (clipped). However, across the range of frames, any particular pixel location will be have one or more well-exposed (lucky) samples that can be combined; 
and different frames will contribute lucky samples in different regions.
%Varying $A_i$ and $\Delta t_i$ aims to find frames $V_i$ such that for any particular pixel location, that are simultaneously not too noisy and not too clipped.
%
Note that this formulation is only valid in the absence of camera or scene motion, so
\sects{sec:align} and~\ref{sec:merge} describe how we align the scene content and then determine weights $w_i$ for fusing the frames.
%
%works when the frames are perfectly aligned and there is no scene motion, it is incorrect when motion is not controlled. %Thus, alignment of frames is necessary and we describe our approach in the next section.
% In the following sections, we describe how to capture, align, and set the weights $w_i$ for fusion.
% In the following, we describe the capture procedure for this burst -- how we set all $A_i, \Delta t_i$ -- before describing the motion-aware alignment and merging.
\vspace{-3mm}
\subsection{Capture Procedure}
\label{sec:capture}

Our capture procedure is controlled by an automatic exposure (AE) algorithm consisting of two parts: 
\emph{viewfinding}, which runs in a continuous loop to produce an acceptable preview; and 
\emph{bracket capture}, which computes the best sequence of settings to use given its most recent knowledge of the scene.
Viewfinder AE must run in real-time on a smartphone.
Therefore, we optimize a simple loss balancing noise (dominated by the shadows) and highlight clipping.
Given an input raw frame $V$, we first compute its $15$-th percentile intensity $x$.
Our calibrated affine noise model $\sigma^2(x) = ax + b$ lets us compute the SNR of the shadows as $S = x / \sigma(x)$~\cite{hasinoff2010noise}.
Second, we compute a soft clipping score $C$, where intensities between $0.9$ and $1$ transition from ``not clipped'' to ``clipped'' via a sigmoid.
The loss is then $L(V) = C(V) - \lambda\ \mathrm{log}\ S(V)$ and we choose the \emph{total exposure} (duration times gain) that minimizes $L$.
In the absence of clipping, we can estimate $L$ for any exposure by scaling $V$: an underexposed scene can be driven to a desired level of noise in the shadows, while clipping some fraction of pixels.
If there is significant clipping, we divide the total exposure by 2 until the scene is once again underexposed.

% We capture a burst of frames $V_i(A_i, \Delta t_i) \; \forall i \in \{1,\ldots, n\}$ metered by an automatic exposure (AE) algorithm.
% Before shutter press, the system runs a \emph{continuous} control loop in order to drive the viewfinder. 
% Given an input raw frame, AE computes the duration and gain (the \emph{total exposure}) of the next frame.
% As it must run in real-time on a smartphone, we optimize a loss balancing noise (dominated by the shadows) and highlight clipping.
% From the linear raw image $V$, we first compute its $15$-th percentile intensity $x$.
% Our calibrated affine noise model $\sigma^2(x) = ax + b$ lets us compute the SNR of the shadows as $SNR = x / \sigma(x)$~\cite{hasinoff2010noise}.
% Second, we compute a soft clipping score $C$, where intensities between $0.9$ and $1$ transition from ``not clipped'' to ``clipped'' via a smooth (sigmoid) function.
% The loss is then $L(V) = SNR(V) + \lambda C(V)$.
% In the absence of clipping, we can estimate $L$ for any exposure by scaling $V$: an underexposed scene can be driven to a desired level of noise in the shadows, while clipping some fraction of pixels.
% We choose the total exposure that minimizes $L$.
% If there is significant clipping, we divide the total exposure by 2 until the scene is once again underexposed. We assume the photographer is holding still and factorize the total exposure that prefers longer exposures and minimum ISO: only when the exposure time exceeds a threshold (1 second) do we increase gain.

We tune $\lambda$ so that at shutter press, the most recent viewfinder settings can be used as an exposure reference:
even without bracketing, it constitutes a usable (albeit slightly dark) photo. %, and a reasonable starting point.
In fully-automatic mode, we capture 5 frames from -2 to +2 stops around the reference, clamping to device limits as well as our own -- no less than 20 dB in the shadows of the shortest frame, and no longer than 1 second exposure since the camera is handheld.
%
% Users can override the limits with manual controls.
Finally, we assume the photographer tries to remain still, and factorize the total exposure so as to prefer longer exposures with minimum ISO: only when the exposure would exceed our 1 second threshold do we increase gain.

% As defined above, we deliberately clip the shortest frame in a minute fraction of pixels for reduced noise, see Supplemental Material. We compute the settings for the rest of the sequence $\Delta t_i$ by \emph{stopping down} to shorter exposures with a uniform step size. Since we have no knowledge of scene content when pixels are clipped, we solve for step size using total capture time.
% Total capture time ranges from 5 seconds for a typical scene and up to 30 seconds for very dark scenes.
%\adam{This is confusing. --> fixed hopefully}
%Our acquisition strategy biases our downstream pipeline towards one gradually adding 

In principle, our method supports an arbitrary capture order, base frame selection, and merge order.
We find the best results if we capture from short to long, with the shortest serving as the base frame, as it typically has the least blur and preserves the most highlight detail. 
The intuition is that it is easier to start with a sharp (albeit noisy) base, and denoise using subsequent frames, rather than starting with a blurry base and attempting to deblur and inpaint highlights. See Supplemental Materials for more details.

\vspace{-4pt}
\subsection{Alignment}
\label{sec:align}

\newcommand{\IT}{\textbf{I}_{\textbf{T}}}
\newcommand{\IB}{\textbf{I}_{\textbf{B}}}
\newcommand{\homography}{\textbf{H}}

\begin{figure}[t!]
\vspace*{-0pt}
    \centering
    \begin{minipage}[b]{0.32\linewidth}
        \centering
        \includegraphics[width=\linewidth]{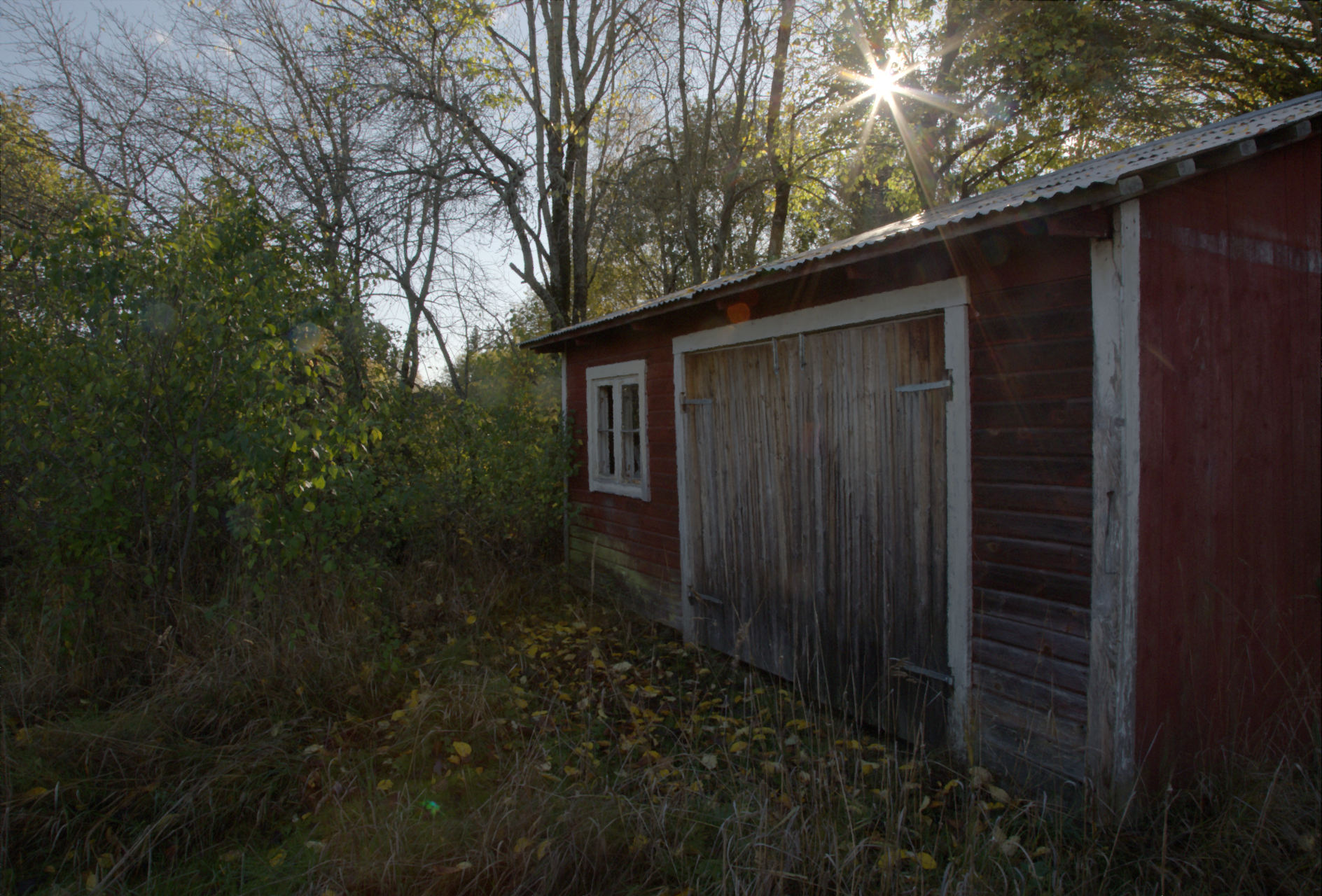}
                \vspace{-8mm}         \caption*{(a) LuckyHDR Output}
    \end{minipage}
    \begin{minipage}[b]{0.32\linewidth}
        \centering
        \includegraphics[width=\linewidth]{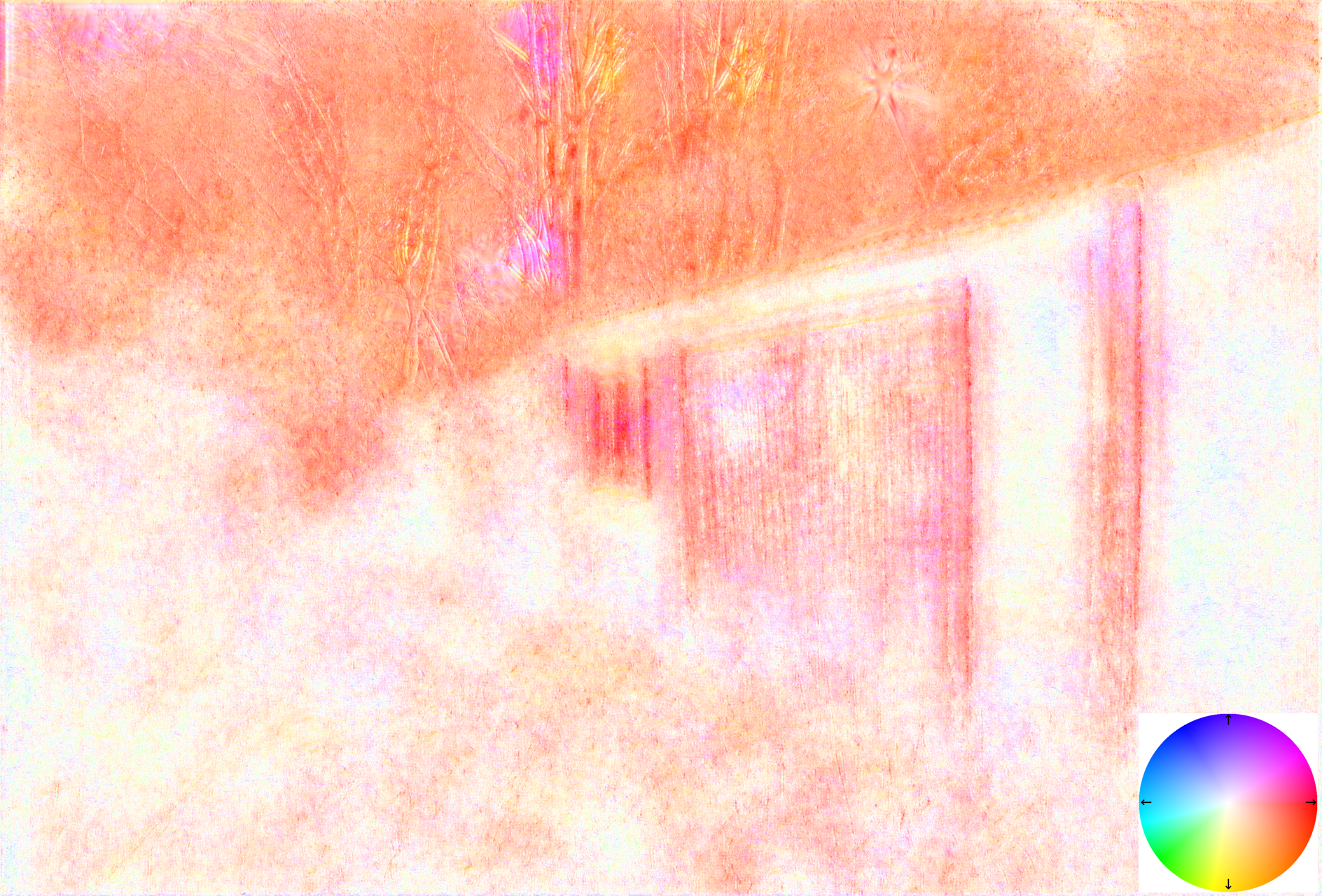}
                \vspace{-8mm}         \caption*{(b) Shift Map Mid}
    \end{minipage}
    \begin{minipage}[b]{0.32\linewidth}
        \centering
        \includegraphics[width=\linewidth]{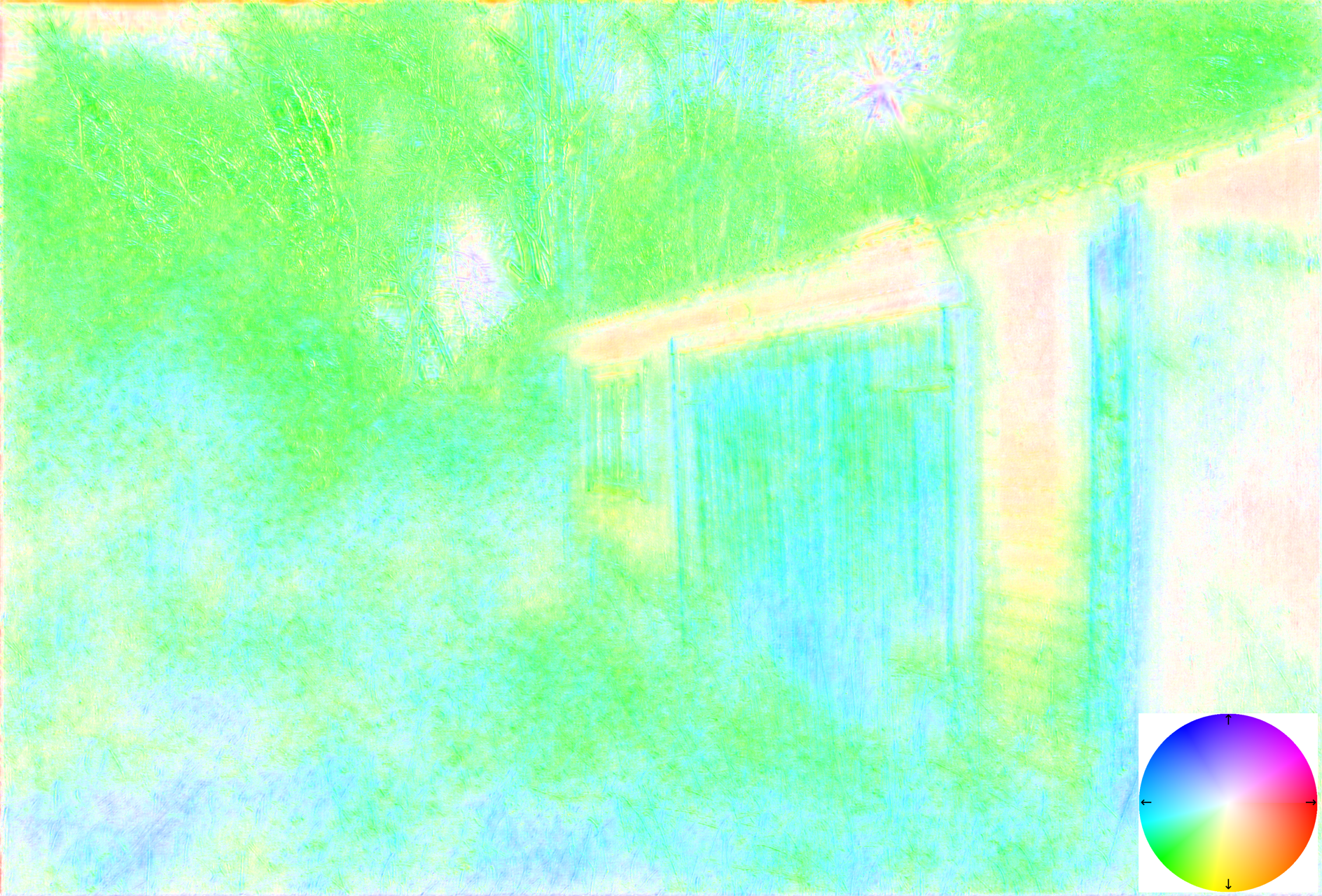}
                \vspace{-8mm}         \caption*{(c) Shift Map Long}
    \end{minipage}

    \begin{minipage}[b]{0.32\linewidth}
        \centering
        \includegraphics[width=\linewidth]{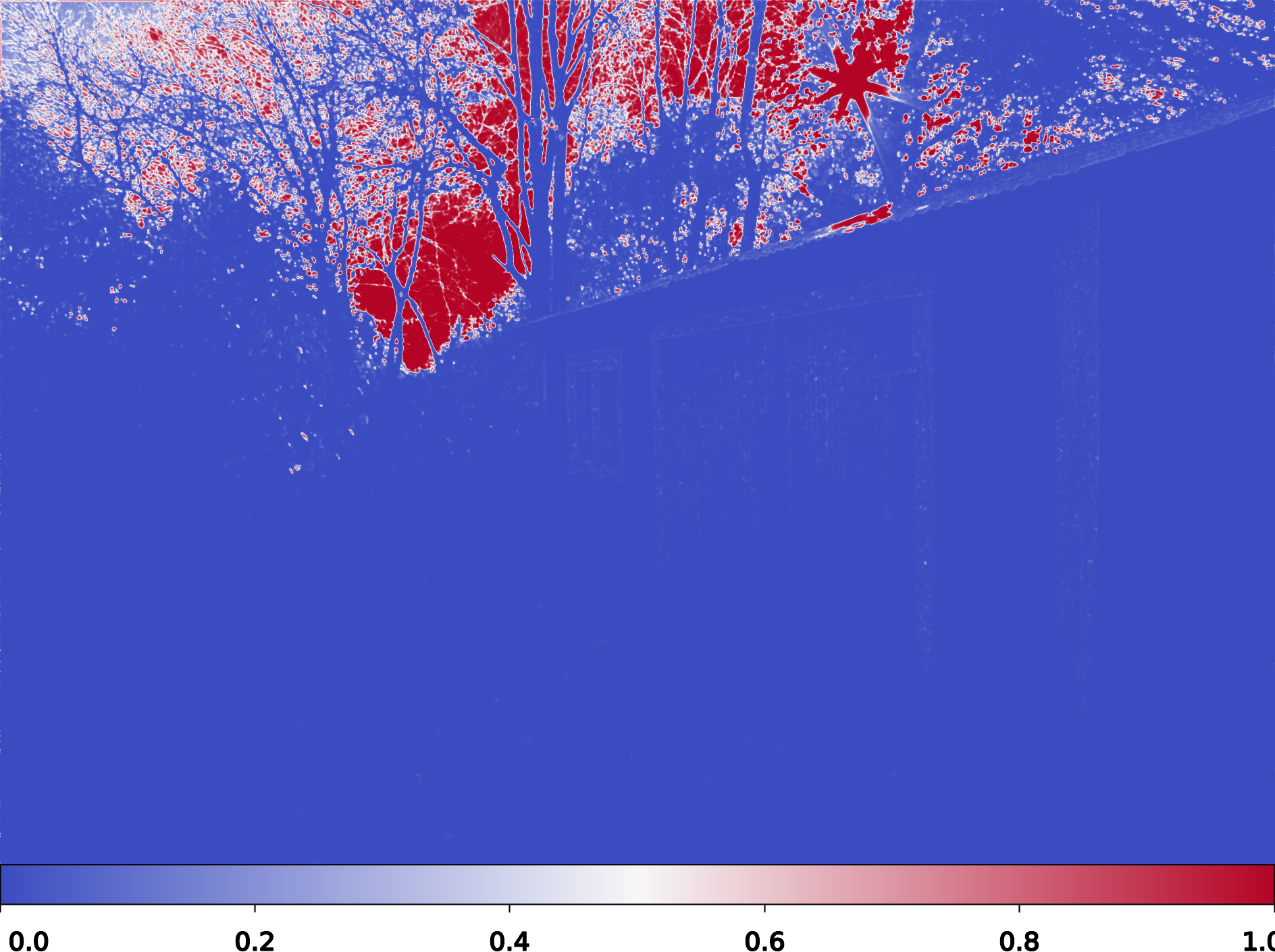}
                \vspace{-8mm}         \caption*{(d) Merge Weights Short}
    \end{minipage}
    \begin{minipage}[b]{0.32\linewidth}
        \centering
        \includegraphics[width=\linewidth]{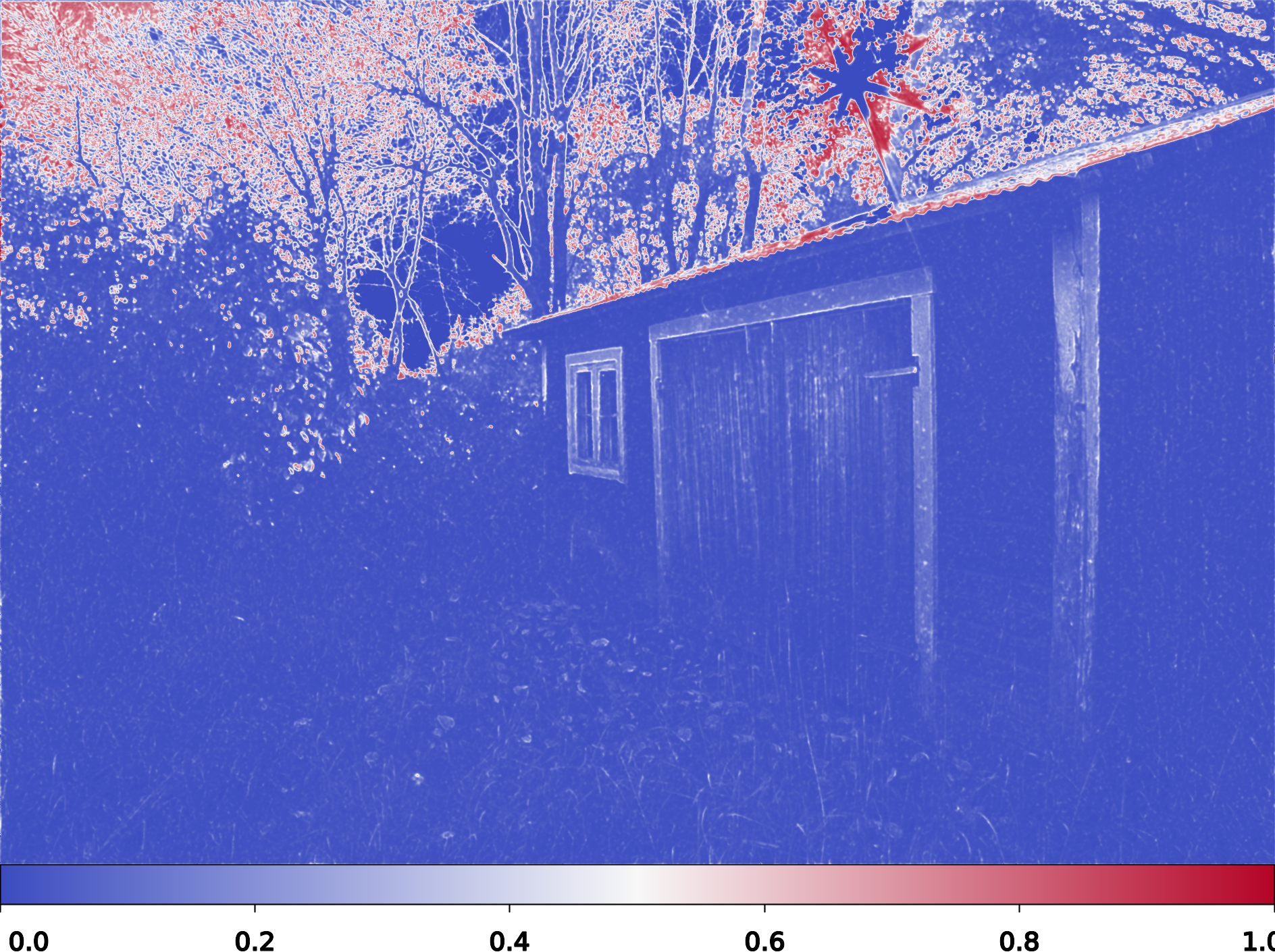}
                \vspace{-8mm}         \caption*{(e) Merge Map Mid}
    \end{minipage}
    \begin{minipage}[b]{0.32\linewidth}
        \centering
        \includegraphics[width=\linewidth]{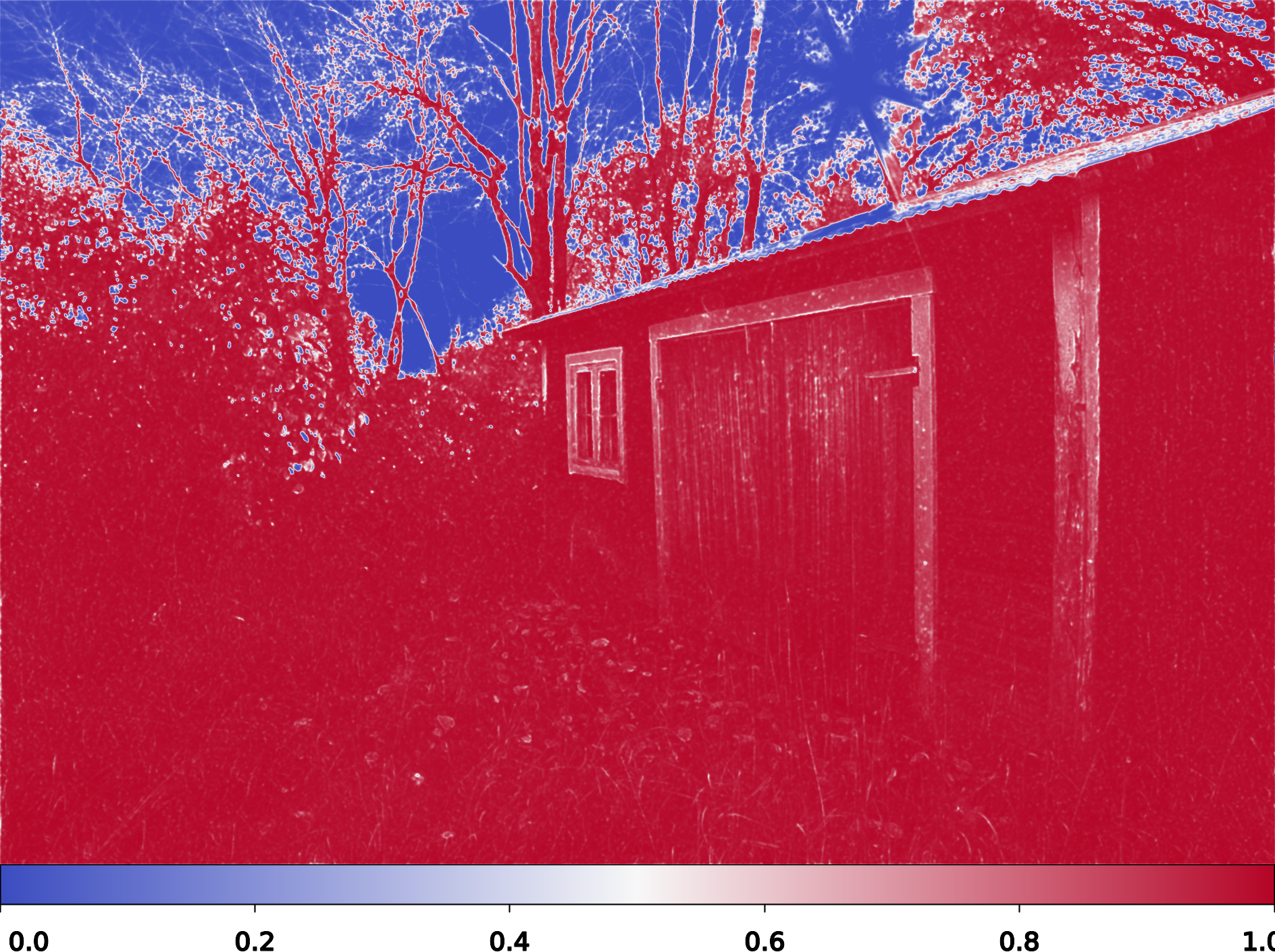}
                \vspace{-8mm}         \caption*{(f) Merge Map Long}
    \end{minipage}\vspace*{-8pt}
    \caption{\textbf{Shift and Merge Weights.} 
    Three frames with short, medium, and long exposures, respectively, are combined by our method, producing (a).
    The short frame is the base and has no shift map. 
    (b-c) Shift maps mid and long are mostly red and green, denoting different shift directions due to shake.
    (d-f) After two iterations, the brightest content is selected from the shortest exposure, and the remainder is filled in mostly from the long exposure. Some regions in the sky are filled in from the mid exposure. The weights (d-f) sum to 1.
    % (d-e) After the first merge iteration, the brightest content is selected from the short exposure, and the remainder is filled in from the medium exposure (weights $d+e=1$).
    % %
    % (f) The second merge iteration replaces shadows and midtones using content from the long exposure -- by multiplying it by the weights in $f$, and then adding $(1-f)\times$ the intermediate result. 
    }
    \Description{Visualization of intermediate outputs: the final LuckyHDR reconstruction, estimated per-frame shift maps, and merge weight maps across two merge iterations for a 3-frame bracket.} 
    \label{fig:ablation_shift_merge}
    \vspace*{-8pt}
\end{figure}

Any burst photography system must account for motion. Even with static scenes and cooperative subjects, the leaves of a tree can sway from a gentle breeze and hands can shake when the photographer breathes.
While the alignment of an image sequence captured with uniform exposure is a well-studied problem~\cite{liba2019verylow,lucas1981opticalflow} -- typically aligning the burst frames sequentially to a single base frame -- the proposed method requires aligning a capture stack with non-uniform exposure. 
% When dealing with bracketed images, we choose the image with the longest exposure that remains free of blur, as it is anticipated to capture the most signal across various regions within the scene. 
We start from the base frame (the shortest exposure frame as chosen by AE as described above) and apply the same alignment-and-merge block sequentially: at each step we treat the current fused estimate as the new ``base frame'' and align the next frame to it before merging.

The alignment and merge networks in our method operate on 4-channel tensors as inputs composed of the exposure-normalized linear RGB concatenated with a tone-mapped luma channel.
%We describe the exposure normalization of this input tensor next before introducing our alignment network. %\jc{I can update this section. How do you compute luma and what's the tone map operator?}

\relatedpar{Exposure Normalization} We normalize the bracket to a common reference exposure (that of the base frame) to bring all frames into a comparable numeric range.
Raw images typically contain ``affine data'' in 16-bit integer format between a sensor black and white level. We subtract black, divide by the effective range (white$-$black), apply a lightweight highlight recovery procedure, and map to linear sRGB; we then clamp to $[0,1]$ for numerical stability, yielding our linear images $I_i$. Next, to compress intensities, we use a global tone curve $T_\mu(x)=\log(1+\mu x)/\log(1+\mu)$~\cite{kong2025safnet}. 

% More sophisticated local motion models such as local mesh warps did not make a significant difference.

%JC: @Ruyu: use sRGB for the final version and re-run. LightGlue is almost certainly trained on sRGB... or maybe Display-P3 images.
%Many options are possible, but we find that using SuperPoint \cite{detone2018superpoint} features matched by LightGlue \cite{lindenberger2023lightglue} with traditional RANSAC worked well.
%At this stage of the pipeline, we do not resample images but simply store the transforms.
%Ruyu: we did resample the image, should we not?
%JC: Nevermind. I think you'll get better flow if you resample only when you compute the loss, but it's probably minor. You need to 
\relatedpar{Coarse-to-fine Alignment}
Denote the base frame $I_B \in \mathbb{R}^{H \times W \times 3}$ and the alternate frame $I_A \in \mathbb{R}^{H \times W \times 3}$, 
% and their gamma encoded counterparts $\Gamma_B$ and $\Gamma_A$. 
we extract an exposure-invariant alignment feature $\phi(\cdot)$, consisting of the globally tone-mapped RGB and the gradient magnitude of its luma channel $L(\cdot)$, that is
\begin{equation}
\begin{aligned}
\phi(I) &= \textsc{Concat}\Big(\textsc{Norm}(T_\mu(I)),
\, \|\nabla\,L(\textsc{Norm}(T_\mu(I)))\|\Big)\in\mathbb{R}^{H\times W\times 4}.
\end{aligned}
\end{equation}
With the two exposure-invariant features $\phi(I_B)$ and $\phi(I_A)$, we form the input to our alignment network by concatenating them and their difference; that is
\begin{equation}
\begin{aligned}
\textsc{input}_{\text{align}} &= \textsc{Concat}\big(\phi(I_A),\,\phi(I_B),\,(\phi(I_A)-\phi(I_B))\big) \in\mathbb{R}^{H\times W\times 12}.
\end{aligned}
\end{equation}
We predict a dense motion map in a coarse-to-fine manner. Let $d$ be the coarse downsampling factor. We first predict a bounded coarse shift at resolution $(H/d,W/d)$, upsample it to full resolution, then predict a bounded residual refinement as
\begin{equation}\label{eq:alignment}
\begin{aligned}
s_c &= d\cdot \textsc{Up}(s_c^{\downarrow}), \;\;\text{with} \;\; s_c^{\downarrow} = m_c\cdot\tanh\!\Big(g_{\theta}^{\textsc{coarse}}\big(\textsc{Down}(\textsc{input}_{\text{align}})\big)\Big),\\
s_f &= m_r\cdot\tanh\!\Big(g_{\theta}^{\textsc{fine}}\big(\textsc{Concat}(\phi(I_A^{c}),\phi(I_B), \phi(I_A^{c})-\phi(I_B))\big)\Big),\\
I_{A}^\triangle &=\mathcal{W}(I_A, s_c + s_f),
\end{aligned}
\end{equation}
where $g_{\theta}^{\textsc{coarse}}$ and $g_{\theta}^{\textsc{fine}}$ are small convolutional networks responsible for coarse and fine alignment, respectively, and $\mathcal{W}$ is bilinear warp and resampling. See the supplement for details. In our default implementation, $d{=}4$ and $m_c{=}13$ (pixels at coarse resolution), so the coarse stage $g_{\theta}^{\textsc{coarse}}$ can tolerate up to $d\!\cdot\!m_c \approx 52$ pixels at full resolution. We set $m_f{=}6$ pixels for the residual refinement stage $g_{\theta}^{\textsc{fine}}$. For the fine stage, we parameterize shift predictions via $\tanh(\cdot)$ and fixed scaling to keep gradients stable and to reflect the small-motion regime expected after exposure normalization.

\begin{figure*}[t!]
\centering\includegraphics[width=0.99\textwidth]{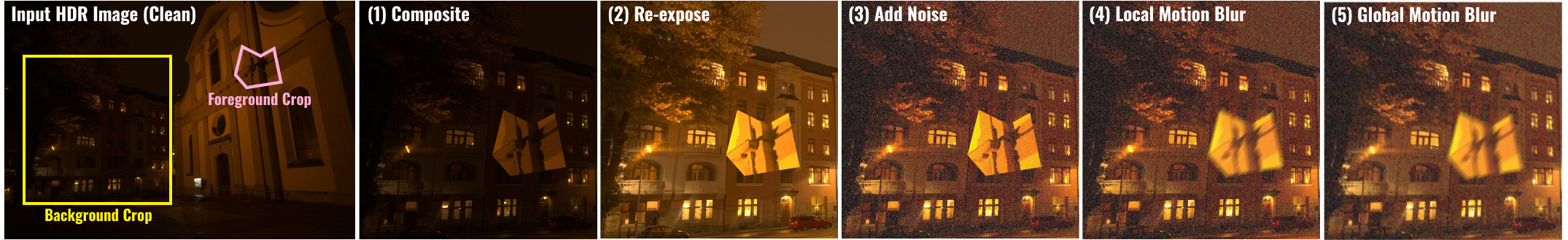}
    \vspace{-3mm}
    \caption{\textbf{Bracketed Burst Simulation.} \rev{We generate realistic training data by starting from a clean HDR image and (1) synthesize local motion by cropping a foreground patch with a random polygon mask and compositing it onto a translated background, (2) re-expose the composite using simulated exposure scales, (3) add spatially varying shot and read noise, and (4–5) apply motion blur to the foreground and background, simulating local motion and handheld camera shake. See text in Section~\ref{sec:data} for details.}}
    \label{fig:motion-syn}
    \vspace{-1mm}
\end{figure*}
\vspace{-3mm}
\subsection{Merging}
\label{sec:merge}

%For alignment we derive a 4-channel exposure-invariant feature $\phi(\cdot)$ from the linear RGB, and for merging we use a lightweight 4-channel feature that appends a gamma-space luma channel to linear RGB.}

After warping, we obtain linear $I_{A}^\triangle$ from Eq.~\eqref{eq:alignment}
% gamma-encoded $\Gamma_{A}^\triangle$ versions 
of the alternate frame in base frame coordinates, and aim to combine these two frames into an output with reduced noise and increased dynamic range.
Rather than directly synthesizing output image pixels, we instead opt for a lightweight approach that predicts at each pixel blending weight in $[0, 1]$.

\begin{figure*}
\vspace{-11pt}
    \centering
\includegraphics[width=1.0\linewidth]{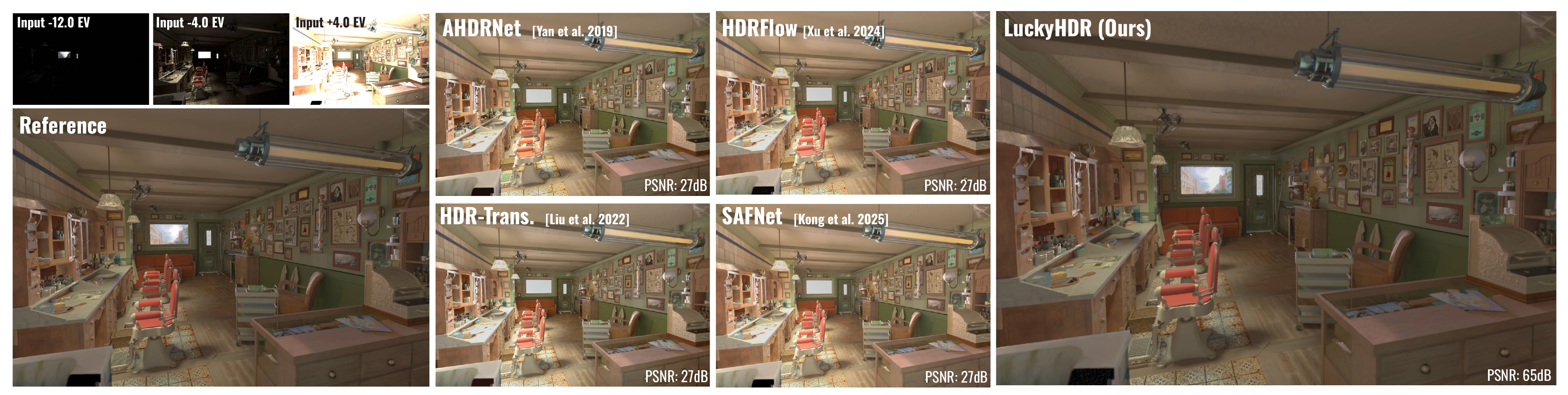}
\vspace*{-5mm}
    \caption{\textbf{Extreme High Dynamic Range.} Our iterative approach can even handle this extreme 9-frame bracketing sequence, rendered in Blender with simulated hand shake. The inputs are shown in linear space so while the outdoor sky is visible in the -12 EV capture, the rest of the scene is essentially black. At the other extreme +4 EV, mid-tones on the right wall are well captured but the window and much of the left side are blown out. The LuckyHDR reconstruction captures details throughout the scene, though tone mapping struggles to visualize this full range. PSNR shown in lower-right.}
    \Description{An extreme HDR example showing a barbershop example rendered using Blender as a 9-frame exposure bracket from -12 to +4 EV, and the resulting LuckyHDR reconstruction recovering both highlight and shadow details, compared with baseline methods.}
    \label{fig:barbershop}
\vspace*{-2mm}
\end{figure*}

% from the moon:
% The inputs are shown in linear space so while the moon is visible in the -7 EV capture, the rest of the scene is essentially black. At the other extreme +7 EV, mid-tones are well captured but the moon and its reflection in the water are blown out.
% The LuckyHDR reconstruction captures details in the moon as well as the bridge and city in the distance.
% The tone mapped LuckyHDR image preserves details including waves in the water and lights of the moving cars (despite streaks caused by 8 sec exposure at +7 EV). Careful examination reveals faint stars near the moon.
    
% From main text: To demonstrate the robustness of the iterative align-and-merge approach we test it on a synthetic (rendered) scene involving an extreme 9-frame bracketing sequence, with simulated hand shake, straddling more than 20 camera stops, shown in \fig{fig:barbershop}. Our method is able to recover detail across the entire dynamic range (though tone mapping struggles to visualize this range), while baseline methods clip throughout the window area. Note that our current mobile capture app is be unable to capture such inputs because of the extreme range of exposures, but in principle such an app could be built and would require a patient and steady photographer!

The merge stage takes as input both unnormalized intensity signal and a lightweight brightness cue (gamma-space luma channel). Let $\Gamma(I)=I^{1/\gamma}$ and $L(\Gamma(I))$ be its luma channel. We extract a per-frame merge feature
\begin{equation}
\psi(I)=\textsc{Concat}\big(I,\,L(\Gamma(I))\big)\in\mathbb{R}^{H\times W\times 4},
\end{equation}
and then concatenate the base and warped alternate features as the input to the merge network
\begin{equation}
\textsc{input}_{\text{merge}} = \textsc{Concat}\Big(\psi(I_B),\, \psi(I_A^{\triangle})\Big)\in\mathbb{R}^{H\times W\times 8}.
\end{equation}
The merge weight predictor $f_{\theta}^{\textsc{merge}}$ is a lightweight three-layer CNN that predicts two-channel logits, followed by a softmax to obtain convex weight
\begin{equation}
\begin{aligned}
w_\text{logit} &= f_{\theta}^{\textsc{merge}}\!\left(\textsc{input}_{\text{merge}}\right),\\
(w_B, w_A) &= \textsc{Softmax}\!\left(w_\text{logit},\, \text{dim}=1\right).
\end{aligned}
\end{equation}
We finally fuse in linear space as a per-pixel convex combination
\begin{equation}
I_\textsc{merged} = w_B \odot I_B + w_A \odot I_A^{\triangle}.
\end{equation}
In practice, we additionally gate $w_A$ by the warp validity mask (out-of-bounds pixels from bilinear resampling) and renormalize $(w_B,w_A)$ to sum to $1$, ensuring that the output remains a convex combination of valid input pixels. As the weights are produced by a softmax, they are non-negative, and this property is preserved by the gating and renormalization.
%
% We do not impose any type of intermediate supervision or regularization on the shift maps $\mathcal   {G}$ or weights $w$ and rely exclusively on end-to-end learning to determine the weights.->BL: It is not the case
% To penalize alignment artifacts, we perform intermediate supervision of the merge block by with the $\ell_1$ loss between the predicted shifted and known shifted image.

\relatedpar{Iterative Align and Merge}
By design, a single merge step results in a linear combination of linear images, where the weight $w$ at each pixel is learned.
Repeated application (as in \fig{fig:luckyhdr-overview}) is equivalent to iterative alpha compositing with learned per-pixel alphas.
%We note that the final output remains a \emph{convex} combination of the input images - it is linear in the inputs but does not have linear precision in the weights (so they are not barycentric coordinates).
This formulation lets us keep the network lightweight, supporting interactive rates at high resolution on a smartphone.

We find that the system is able to optimize for suitable weights using only our synthetic dataset described in the following section.
It is able to cope with noise in deep shadows, spatially varying highlight clipping, and blur.
\fig{fig:ablation_shift_merge} shows how the network learns to exclude clipped pixels (by setting the weights to nearly zero).

\vspace{-0.25\baselineskip}
\section{Dataset and Training}
\label{sec:data}

To train our reconstruction pipeline, we generate a synthetic dataset that mimics realistic handheld acquisition. 
Note that we focus only on creating the degradations targeted by the trainable modules of our pipeline; i.e., small local motion and bracketed exposures with spatially varying noise.
% We do not attempt to model 
We propose a method to accurately simulate handheld exposure bracketing \rev{as illustrated in Figure \ref{fig:motion-syn}}. Without loss of generality, we apply our motion simulation on the SI-HDR dataset~\cite{hanji2022sihdr}. 

% Exposure and noise
To generate synthetic capture stacks of $N$ frames, we start with a single clean HDR image and re-expose it with $e_i, i\in0 \ldots N-1$ exposure scales \rev{as in Figure \ref{fig:motion-syn} (2)}. Images in the SI-HDR dataset are first preprocessed to linear sRGB space. We normalize each HDR image by its 99.9th percentile and clip to $[0,1]$, then apply a random power-law augmentation with exponent sampled from $[1,1.5]$ to form a well-exposed image $H$. Then, we re-expose this image by approximating Eq.~\eqref{eq:img-formation} following Hanji~\etal~\shortcite{hanji2022sihdr}, as
\vspace{-2pt}
\[I_i=\text{round}(\min(e_iH+\eta(e_iH), 1)), \quad \eta\sim\mathcal{N}(0, n_s x + n_o),
\]
where camera noise $\eta$ is modeled with the shot noise $n_s$ and the read noise $n_o$ \cite{hasinoff2010noise}. We assume no prior knowledge of camera noise profile; and therefore, uniformly sample $n_s \in \left[10^{-5},10^{-3}\right]$ and $n_o \in \left[10^{-7},10^{-5}\right]$. \rev{See noise added in Figure \ref{fig:motion-syn}, step (3).}

% Camera motion
We augment the bracketed stack with both local moving objects and global hand shake, typical in handheld photography. For local motion, we add a dynamic foreground layer to each scene, which simulates objects like moving cars and pedestrians. We deliberately expect such motion to be too large to align across the stack, and encourage the model to ignore these misalignments and instead fall back to the base frame. Following Mayer et al.~\shortcite{mayer2018opticalflow}, we crop and translate a patch from the HDR image (before bracketing) using a random polygon mask, then composite it after flipping to reduce visual repetition, \rev{as in Figure \ref{fig:motion-syn}, step (1)}.

For global motion induced by hand shake, we sample per-frame translation offsets $\Delta_i$ (in pixels) and apply them by shifting the bracketed frames. We use a two-component mixture: with probability $p$, we sample from $[-m_l, m_l]^2$, and otherwise from $[-m_s, m_s]^2$ (we use $p=0.05$, $m_s=2$, and $m_l=16$). Our simulator also emits an optional validity mask for analysis and ablation: moving-foreground regions are marked invalid, and frames with large translations (e.g., $>7$~px) or synthetic motion blur are flagged as unmatchable.
To further simulate motion artifacts, we apply blur to both foreground and background separately\rev{, see Figure \ref{fig:motion-syn}, step (4) and (5)}. We apply motion blur with probability $0.3$, scaling blur radius with exposure time (up to 5~px for the background and 7~px for the moving foreground). We do not blur the base frame with shortest exposure. Finally, the foreground and background are composited to form the final motion-augmented image. 
% \rev{\fig{fig:motion-syn} illustrates our simulation approach. At each bracket exposure time, we pick a foreground patch, translate it by a per-frame object offset, and blur it with a direction-aligned kernel whose length tracks the local object motion. We then alpha-composite the warped patch onto the HDR background to form a per-frame moving-foreground image. After compositing, we apply a second global blur to the whole frame, using a kernel whose radius grows with the exposure of that frame, so that longer exposures carry more camera-shake blur than the shortest one. The foreground object motion, the global translations from \sect{sec:data}, and this exposure-tied blur are sampled independently per frame, so a given training stack mixes object motion, parallax-free hand shake, and exposure-dependent blur in the same way a real handheld burst would. See the Supplemental Material for further implementation details.}

% \vspace{-5mm}
\vspace*{-0.25\baselineskip}
\subsection{Training}
We train LuckyHDR entirely on synthetic data in two phases (additional details in the supplement): (i) supervised training on the training set described above (small-shift regime), followed by (ii) an affine-motion curriculum (dominated by translation, with mild zoom) that expands the camera-translation range up to 50~pixels by updating only the coarse alignment module.

\relatedpar{Phase I: Small-motion Training.}
Given a synthetic burst $\{I_i\}_{i=0}^{N-1}$ and clean reference $H$, we supervise the HDR reconstruction with an $\ell_1$ loss in the tone-mapped domain via
\begin{equation}
\begin{aligned}
T_\mu(x) &= \frac{\log(1+\mu\,\max(x,0))}{\log(1+\mu)}, \quad
\mathcal{L}_{\text{pred}} = \big\|T_\mu(\hat{H}) - T_\mu(H)\big\|_1.
\end{aligned}
\end{equation}
This balances errors across the entire dynamic range. To explicitly guide alignment without requiring ground-truth flow, our simulator also produces a ``no-shift'' target for each frame (same exposure/noise/blur, but without the global jitter). During training, we form a validity mask $M_i$ that defines warp validity (out-of-bounds pixels from bilinear sampling) and dataset-specific masks (e.g., extreme blur/unmatchable regions). For the results reported in this paper, we rely on the validity mask and learned merging rather than hard validity gating. We apply intermediate supervision on the alignment stage by comparing each warped alternate frame to its corresponding no-shift target in exposure-normalized linear RGB via
\begin{equation}
\mathcal{L}_{\text{warp}} = \sum_{i\neq B}\big\|M_i \odot (I_i^{\triangle} - I_i^{\text{no-shift}})\big\|_1,
\end{equation}
where $B$ is the base-frame index, $I_i^{\triangle}$ is the model-warped frame, and $M_i$ is the validity mask. For translation-only motion, we also penalize the spatial variance of the predicted shift fields to discourage high-frequency warps that can distort texture; that is
\begin{equation}
\mathcal{L}_{\text{var}} = \sum_{i\neq B} \mathrm{Var}\big(\Delta_i\big),
\end{equation}
and we use a weighted sum of $\mathcal{L}_{\text{pred}}$, $\mathcal{L}_{\text{warp}}$, and $\mathcal{L}_{\text{var}}$. We learn a single set of weights that is used for all iterations.

\relatedpar{Phase II: Large-motion Curriculum Training}
To learn large motion alignment, we start with the small motion checkpoint and train only the coarse alignment module on an affine-motion variant of the simulator that applies global camera motion after bracketing (translation with mild zoom; rotation disabled). We mix affine-motion batches with the Phase-I small-motion simulator as a rehearsal, and gradually increase the translation range from 0--14~px to 21--50~px with an increasing affine sampling ratio. 
% We select the final checkpoint by maximizing large-motion validation quality while constraining the small-motion PSNR$_\mu$ drop to be within 0.1~dB (per-step) relative to the rehearsal baseline. We train with Adam~\cite{kingma2014adam} on $512\times512$ patches and use a cosine learning-rate schedule.}

%\vspace{5pt}
   
\vspace{-3mm}
\section{Evaluation}
\label{sec:evaluation}

In this section, we compare LuckyHDR against several recent baseline methods.
On synthetic data with ground truth, we conduct quantitative comparisons.
We then assess the contribution of different components of our design with a set of ablations.
Finally, we conduct a qualitative comparison to assess generalization on real photographs from several cameras -- captures from all cameras are unseen to the methods evaluated. 
%held-out samples from the synthetic dataset described in the previous section and on experimental smartphone data that we acquired with our iPhone capture application described in Sec.~\ref{sec:capture}. Next, we first validate the design choices of our method with ablation experiments before reporting experimental findings.

\begin{table}[t]
\vspace{-11pt}
  % Generated from SIHDR-fast eval JSONs (seed=46):
  %   python hdrproj/1211/LuckyHDR/tools/make_quan_table_tex.py --step 2 --macro bl
  %   python hdrproj/1211/LuckyHDR/tools/make_quan_table_tex.py --step 3 --macro bl
  % ITPI (A6000, 1280x1888, 3 inputs): hdrproj/1211/LuckyHDR/exp/itpi_sihdr_fast_a6000_job26229178.json
  %   python hdrproj/1211/LuckyHDR/tools/benchmark_rescost_a6000.py --only_hw --hw 1280 1888 --warmup_hw 5 --runs_hw 20
  % HDR+ (py) JSONs (seed=46):
  %   metrics+HDR-VDP2: hdrproj/1211/LuckyHDR/exp/baseline_hdrplus_py_sihdr_fast_test_seed46_job26170625.json
  %   LPIPS:           hdrproj/1211/LuckyHDR/exp/lpips_hdrplus_py_sihdr_fast_test_seed46_job26181322.json
  % HDRFlow/p JSON (seed=46):
  %   step=2/3 test split: hdrproj/1211/LuckyHDR/exp/baseline_hdrflow_official2e_sihdr_fast_test_seed46_job26350384.json
  % AFUNet JSON (seed=46):
  %   step=2/3 test split: hdrproj/1211/LuckyHDR/exp/baseline_afunet_sihdr_fast_test_seed46_job26351436.json
  % AFUNet ITPI (A6000, 1280x1888, 3 inputs):
  %   hdrproj/1211/LuckyHDR/exp/itpi_afunet_sihdr_fast_a6000_job26353884.json
  % Affine translation-only (0--20 px) eval JSONs (seed=46, patch=512):
  %   step=2: hdrproj/1211/LuckyHDR/exp/motion_ranges_*_seed46_step2_cpu_hv.json
  %   step=3: hdrproj/1211/LuckyHDR/exp/motion_ranges_*_seed46_job26203074_hv.json
  %   HDRFlow/pal 2E) row: hdrproj/1211/LuckyHDR/exp/motion_ranges_hdrflow_official2e_seed46_job26350640_hv.json
  %   HDR+ (py) row:   hdrproj/1211/LuckyHDR/exp/motion_ranges_hdrplus_py_seed46_job26203074_hv.json
  \caption{\textbf{
  Quantitative Reconstructive Quality and Runtime}. We evaluate LuckyHDR and existing state-of-the-art reconstruction methods on the synthetic dataset described in \sect{sec:data}. We select the frame with the shortest exposure time as the base frame. We report inference time per output image (ITPI; lower is better) benchmarked on an RTX A6000 at $1888 \times 1280$. We report LPIPS~\shortcite{zhang2018unreasonable} (AlexNet backbone; lower is better) computed on $\mu$-tonemapped images as a perceptual similarity metric. We also report HDR-VDP2 (quality score; higher is better) under a fixed viewing condition (30 px/deg) with a 1000~nit peak display model for all methods. Note that for fairness, we \emph{re-train all methods under the same setting} (three inputs, with step size randomly sampled from $\{2,3\}$ during training), and evaluate the inference results under two different settings: three inputs separated by two stops, and three inputs separated by three stops. We also report results on a controlled translation-only affine-motion split with translation magnitude 0--20~px (bottom block).We additionally report HDRFlow/p and SAFNet/p, which uses the authors' officially released checkpoint (trained on their datasets) rather than re-training on ours. This model attains high PSNR$_l$ but substantially worse tone-mapped PSNR$_\mu$ and LPIPS scores in our bracketed setting, indicating a lack of generalization.}
  \centering
  \footnotesize
  \vspace*{-8pt}
  \begingroup
  \setlength{\tabcolsep}{3.0pt}
  \renewcommand{\arraystretch}{1.05}
  \begin{tabular}{lcccccc}
    \toprule
    Model             & ITPI (ms) & Params (K) & PSNR$_{l}$ & PSNR$_\mu$ & HDR-VDP2 & LPIPS \\
    \midrule
    \multicolumn{7}{c}{\textit{Exposure Step Size = 2 EV}} \\
    \midrule
    HDR+~\shortcite{hasinoff2016burst}              & \textbf{8}  & \textbf{0}  & 43.7      & 27.3           & 31.2       & 0.427       \\
    SAFNet~\shortcite{kong2025safnet}            & 208   & 1120   & 44.3      & 30.6           & 31.0       & 0.249       \\
    SAFNet/p~\shortcite{kong2025safnet}            & 208   & 1120   & 36.8      & 25.5           & 27.3       & 0.460       \\
    AHDRNet~\shortcite{yan2019attention}           & 504  & 1520    & 40.4      & 30.0            & 29.8       & 0.305       \\
    HDRFlow~\shortcite{xu2024hdrflow}           & 61   & 3270   & 48.7      & 33.2             & 38.1       & \textbf{0.226}        \\
    HDRFlow/p~\shortcite{xu2024hdrflow}           & 61   & 3270   & \textbf{50.2}      & 26.7             & 32.7       & 0.507        \\
    HDR-Trans.~\shortcite{liu2022ghost}   & 9371  & 1220   & 37.9      & 32.0             & 36.7       & 0.267       \\
    AFUNet~\shortcite{li2025afunet}   & 21318  & 1162   & 38.4      & 27.7             & 35.7       & 0.332       \\
    LuckyHDR          & 62    & 66   & 50.0     & \textbf{33.6}            & \textbf{40.5}       & 0.241       \\
    \midrule
    \multicolumn{7}{c}{\textit{Exposure Step Size = 3~EV}} \\
    \midrule
    HDR+~\shortcite{hasinoff2016burst}              & \textbf{8}  & \textbf{0}    & 44.1          & 28.4                 & 30.6          & 0.338           \\
    SAFNet~\shortcite{kong2025safnet}            & 208     & 1120     & 46.4          & 32.4                  & 33.8          & 0.164           \\
    SAFNet/p~\shortcite{kong2025safnet}            & 208     & 1120     & 34.2          & 25.4                  & 26.6          & 0.465           \\
    AHDRNet~\shortcite{yan2019attention}           & 504    & 1520      & 40.4          & 32.2                  & 30.0          & 0.305         \\
    HDRFlow~\shortcite{xu2024hdrflow}          & 61      & 3270    & 47.9          & 35.0                  & 38.6          & 0.120           \\
    HDRFlow/p~\shortcite{xu2024hdrflow}          & 61      & 3270    & 49.9          & 26.1                  & 31.4          & 0.535           \\
    HDR-Trans.~\shortcite{liu2022ghost}   & 9371    & 1220     & 37.9          & 31.8                & 36.4          &  0.253          \\
    AFUNet~\shortcite{li2025afunet}   & 21318  & 1162   & 38.5      & 30.6             & 35.4       & 0.335       \\
    LuckyHDR          & 62     & 66      & \textbf{50.2}         & \textbf{36.5}                   & \textbf{43.7}          & \textbf{0.107}           \\
    \midrule
    \multicolumn{7}{c}{\textit{Affine Translation Only (0--20 px)}} \\
    \midrule
    \multicolumn{7}{c}{\textit{Exposure Step size = 3~EV}} \\
    \midrule
    HDR+~\shortcite{hasinoff2016burst}              & \textbf{8}  & \textbf{0}   & 45.5      & 28.0          & 46.8       & 0.308       \\
    SAFNet~\shortcite{kong2025safnet}            & 208   & 1120   & 42.4      & 29.4            & 45.3       & 0.220       \\
    SAFNet/p~\shortcite{kong2025safnet}            & 208   & 1120   & 38.0      & 24.7            & 38.8       & 0.441       \\
    AHDRNet~\shortcite{yan2019attention}           & 504   & 1520   & 43.7      & 31.4         & 37.4       & 0.311       \\
    HDRFlow~\shortcite{xu2024hdrflow}           & 61   & 3270   & 49.8      & 31.6            & 48.5       & 0.154        \\
    HDRFlow/p~\shortcite{xu2024hdrflow}           & 61   & 3270   & 51.8      & 26.2            & 44.8       & 0.503        \\
    HDR-Trans.~\shortcite{liu2022ghost}   & 9371  & 1220    & 42.7      & 31.9            & 51.0       & 0.270       \\
        AFUNet~\shortcite{li2025afunet}   & 21318  & 1162   & 42.4      & 30.5            & 51.4       & 0.333       \\

    LuckyHDR          & 62   & 66    & \textbf{52.0}     & \textbf{33.0}            & \textbf{53.2}       & \textbf{0.143}       \\
    \bottomrule
  \end{tabular}
  \endgroup
  \label{tab:quan}
  \vspace*{-5pt}
\end{table}

\vspace{-3mm}
\subsection{Synthetic Assessment}
\label{sec:synthetic}
We report model size and inference time per output image (ITPI) in \tbl{tab:quan}. Among the neural baselines, LuckyHDR has only 66K parameters, which is $\sim$50$\times$ fewer than HDRFlow (3.27M) and at least $\sim$17$\times$ fewer than the other deep baselines (1.1--1.5M).

\begin{figure*}[t]
\vspace{-5pt}
\centering\includegraphics[width=0.99\linewidth]{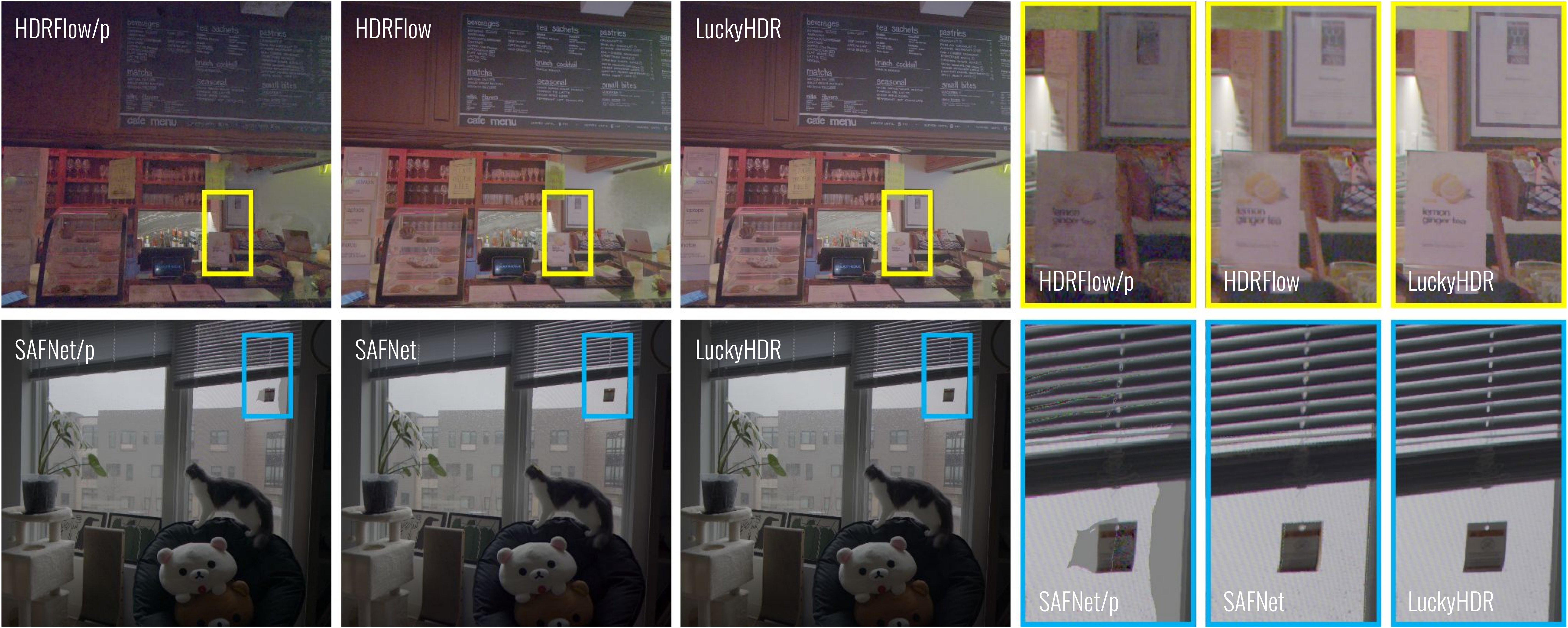}
    \vspace{-8pt}
    \caption{\rev{\textbf{Effect of Simulated Training Data}} We compare two representative real hand-held brackets under \rev{random} hand shake. For each baseline (HDRFlow and SAFNet), we show results from the authors' official checkpoint (HDRFlow/p or SAFNet/p), our variant retrained on our synthetic data, and LuckyHDR. Our dataset improves robustness to our bracketing distribution, but direct pixel prediction still tends to over-smooth fine textures or introduce hallucinated artifacts. LuckyHDR preserves details via explicit alignment and convex fusion of observed pixels.}
    \label{fig:contrib_data}
    \vspace{-8pt}
\end{figure*}

To evaluate the output of the proposed method on simulated data with known ground truth, we measure reconstruction quality in both tone mapped (PSNR$_\mu$) and linear (PSNR$_{l}$) domain and include HDR-VDP2~\cite{mantiuk2011hdrvdp2} as an additional perceptual quality metric under a fixed display/viewing condition.
We do not report SSIM$_{l}$ since all methods achieve values close to 0.99 and not discriminative. 
We also report LPIPS~\cite{zhang2018unreasonable} on $\mu$-tonemapped images as a complementary perceptual similarity metric.
We compare LuckyHDR to SAFNet~\cite{kong2025safnet}, AHDRNet~\cite{yan2019attention}, HDR-Transformer~\cite{liu2022ghost}, AFUNet~\cite{li2025afunet}, HDRFlow~\cite{xu2024hdrflow}, and HDR+~\cite{hasinoff2016burst}. The original HDR+ pipeline targets uniform-exposure burst denoising and relies on camera-specific RAW/DNG processing and display-processed outputs, which makes a fair adaptation to exposure-bracketed HDR evaluation non-trivial. We modify HDR+ to operate on bracketed captures, see Supplemental Material for details. We report results in \tbl{tab:quan}.
To evaluate the \emph{contribution of the proposed architecture in isolation}, we re-train all learning-based baselines on our synthetic training data with three input frames, where the exposure step size is randomly sampled from $\{2,3\}$~EV during training (\sect{sec:capture}). We then evaluate each method for two fixed settings (exposure step size = $2$~EV and $3$~EV). To evaluate the \emph{contribution of the dataset in isolation}, we additionally report HDRFlow/p and SAFNet/p in \tbl{tab:quan} by running the authors' official checkpoints without re-training on our synthetic data. \rev{\fig{fig:contrib_data} depicts two bracketed captures taken in the presence of a handshake with our custom iPhone app. To tease apart the role of architecture and training data for the two strongest baselines, we evaluate HDRFlow and SAFNet in three configurations: the authors' released checkpoint trained on their original datasets (the ``/p'' row in \tbl{tab:quan}), the same architecture retrained from scratch on our synthetic handheld-bracket data, and LuckyHDR. Here, the pre-trained HDRFlow and SAFNet models exhibit heavy ghosting along occluding edges and hallucinated highlights in near-clipped regions. Retraining on our generated training data removes most of these gross artifacts on the same captures, showing that the generated distribution, not the network itself, is what drives generalization to handheld smartphone brackets. Even after retraining, HDRFlow and SAFNet still oversmooth fine texture and occasionally hallucinate sharp edges in bright regions, because they regress pixel values directly. LuckyHDR avoids both failure modes, because each output pixel is a convex combination of pixels observed in the input frames.}

Our quantitative evaluation confirms that LuckyHDR achieves the best overall performance, and remains robust across different step sizes. It also yields the best HDR-VDP2 scores among all compared methods, indicating reduced visibility of artifacts (e.g., ghosting/texture distortions) under a perceptual model of human vision. Relying on estimated optical flow, our approach exploits intensity variation in the input frames, and, as such, prefers larger step sizes. With increasing step size, up to a limit when optical flow estimates are no longer possible to obtain robustly, more information is recovered by the proposed method. Further analysis of different settings is reported in \sect{sec:ablations}.
\begin{figure*}[!p]
\vspace*{-4pt}
    \centering
    \includegraphics[width=0.95\textwidth]{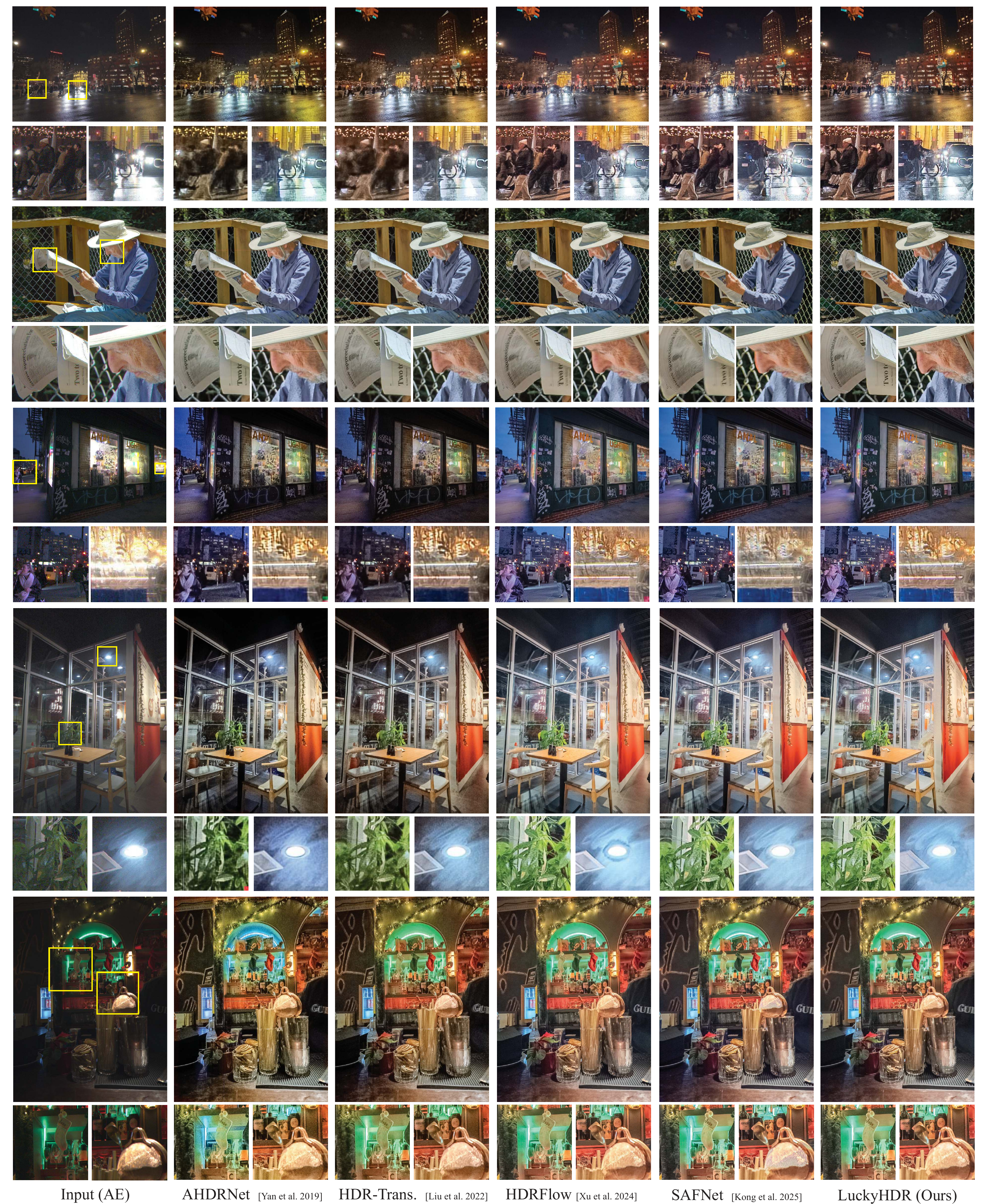}\vspace*{-8pt}
    \caption{\textbf{Handheld HDR Reconstruction Experiments.} LuckyHDR is capable of handling hand shake and local motion. The baseline methods fail to find robust alignment. By comparing the reconstructions with the inputs with different exposure times, we observe that our method simultaneously recovers highlights and shadows without introducing ghosting or noise. See the supplement for more results tone mapped for both SDR and HDR displays.}
	  \Description{A grid of crops from three real handheld HDR brackets, showing two input exposures and reconstructions from multiple baselines versus LuckyHDR.}
	  \label{fig:mobile_comparison}
     %\vspace*{-6pt}
\end{figure*}

\bl{\begin{table}[t]
  % Refreshed from A6000 benchmark JSON (job=26203119):
  %   hdrproj/1211/LuckyHDR/exp/rescost_a6000_job26203119.json
  % HDR+ (py) row from A6000 benchmark JSON (job=26221452):
  %   hdrproj/1211/LuckyHDR/exp/rescost_a6000_job26221452.json
  % AFUNet rows from A6000 benchmark JSON (job=26357725):
  %   hdrproj/1211/LuckyHDR/exp/rescost_a6000_afunet_job26357725.json
  % Generated by: hdrproj/1211/LuckyHDR/scripts/sbatch_benchmark_rescost_a6000_allcs.sh
  \caption{\textbf{GPU Inference Time.} Inference time per image (ITPI) on an RTX A6000 with 50~GB RAM with three input frames.
  We choose the image with shortest exposure as the base frame.
  Dash (-) indicates out of memory on an NVIDIA RTX A6000 GPU with 50~GB memory.}
  \centering
  \footnotesize
  \vspace*{-8pt}
  \begingroup
  \setlength{\tabcolsep}{4pt}
  \renewcommand{\arraystretch}{1.05}
  \begin{tabular}{ccccc}
    Models & Resolution & ITPI (ms) & Params (K) & FLOPs (T) \\
    \midrule
       HDR+& $512 \times 512$ &2 &0 &\textit{n/a}    \\
       SAFNet& $512 \times 512$ &28 &1120 &0.170    \\
       AHDRNet &$512 \times 512$&56&1520&0.379 \\
       HDRFlow& $512 \times 512$ &9 &3270 &0.046    \\
       HDR-Transformer& $512 \times 512$ &1160 &1220 &0.314   \\
       AFUNet& $512 \times 512$ &1543 &1162 &0.302   \\
       LuckyHDR& $512 \times 512$ & 7 &65.8 &0.019  \\
    \midrule
       HDR+& $2048 \times 2048$&13 &0  &\textit{n/a}   \\
       SAFNet& $2048 \times 2048$&367 &1120  &2.73   \\
       AHDRNet &$2048 \times 2048$&892&1520&6.07 \\
       HDRFlow& $2048 \times 2048$&107 &3270  &0.735   \\
       HDR-Transformer& $2048 \times 2048$ &17700 &1220 &5.03   \\
       AFUNet& $2048 \times 2048$&24107 &1162  &4.83   \\
       LuckyHDR& $2048 \times 2048$&107 &65.8 &0.311 \\
    \midrule
       HDR+& $4096 \times 4096$&50 &0  &\textit{n/a}   \\
       SAFNet& $4096 \times 4096$&1500 &1120  &10.9   \\
       AHDRNet &$4096 \times 4096$&-&-&- \\
       HDRFlow& $4096 \times 4096$& 424 &3270  &2.94   \\
       HDR-Transformer& $4096 \times 4096$& -  &- & -    \\
       AFUNet& $4096 \times 4096$&-&-&- \\
       LuckyHDR& $4096 \times 4096$&430 &65.8  &1.24 \\
       % \bottomrule
  \end{tabular}
  \endgroup
\vspace*{-10pt}
  \label{tab:compute}
\end{table}}

\relatedpar{Controlled Translation-only Split} The top of \tbl{tab:quan} reports evaluation results on our SI-HDR-fast handheld simulator (\sect{sec:data}), which includes bracketed exposure/noise, exposure-tied blur, local object motion, and global hand shake with a mixture of small and occasional larger translations. The bottom instead isolates camera translation by using a translation-only affine-motion test split with translation magnitude limited to 0--20~px (no rotation/zoom), so that each method is compared under a controlled global-motion regime. While we do not report >20~px evaluation in the main paper, we still train the coarse alignment module with an affine-motion curriculum up to 50~px (\sect{sec:data}) to better handle occasional large motions in real captures.
%% [Stage 3] figure_iphone_new moved inline from paper_suffix.tex.
\begin{figure*}[t]
\vspace*{-5mm}
    \centering
    \includegraphics[width=1.00\textwidth]{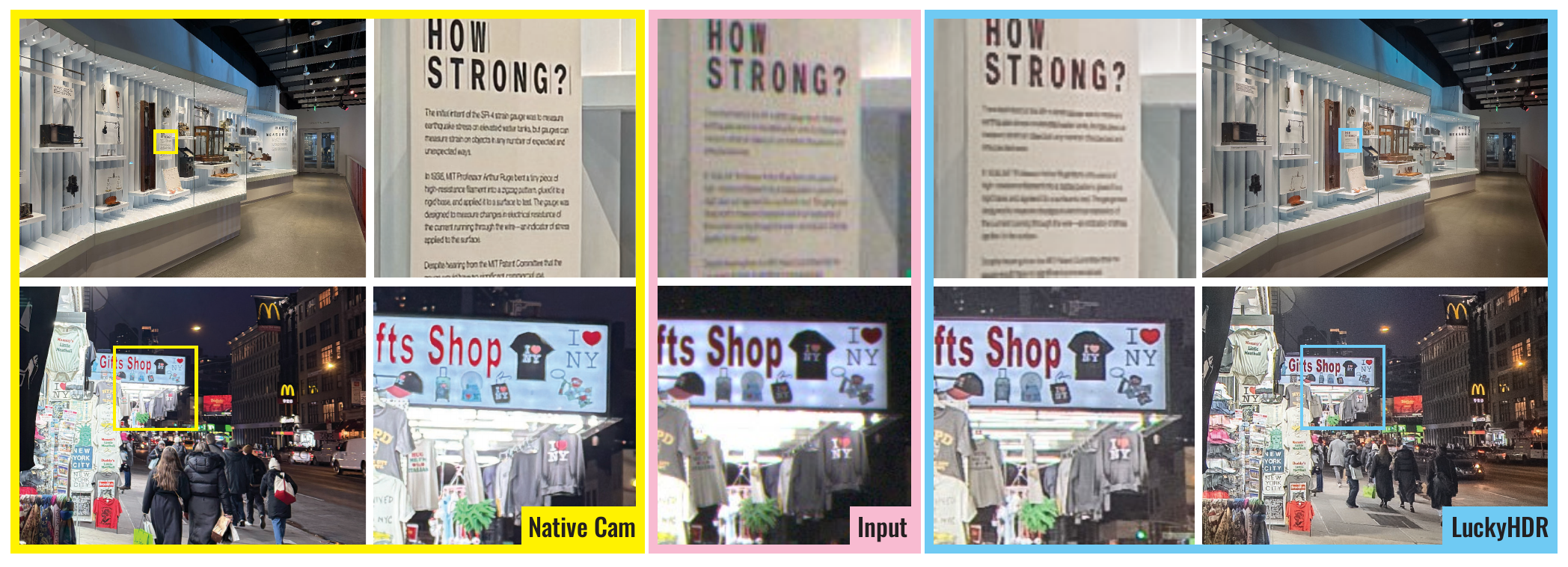}
    \vspace*{-24pt}
    \caption{\rev{\textbf{Comparison vs. Native iPhone Camera.}}
    We photograph the same scenes using both the native iPhone 16 Pro Max camera (left) and our custom app (right, post-processed in Adobe Camera Raw), which follows the acquisition pipeline described in \sect{sec:capture}.
    One of the bracketed input photos is shown in the center.
    Upper row: the iPhone yields sharper lettering in the ``HOW STRONG'' sign, but hallucinates detail in the small text on that sign. Lower row: LuckyHDR more faithfully reproduces the ``Gifts Shop'' sign, as well as the highlights in the fluorescent lighting tubes beneath it.}
    \Description{Comparison of two scenes captured with the native iPhone camera pipeline versus our custom capture app and LuckyHDR processing, trading sharpness for avoiding hallucinations.}
    \label{fig:iphone_comparison}
     \vspace*{-2mm}
\end{figure*}

\begin{figure*}
% \vspace{20pt}
    \centering
\includegraphics[width=1.0\linewidth]{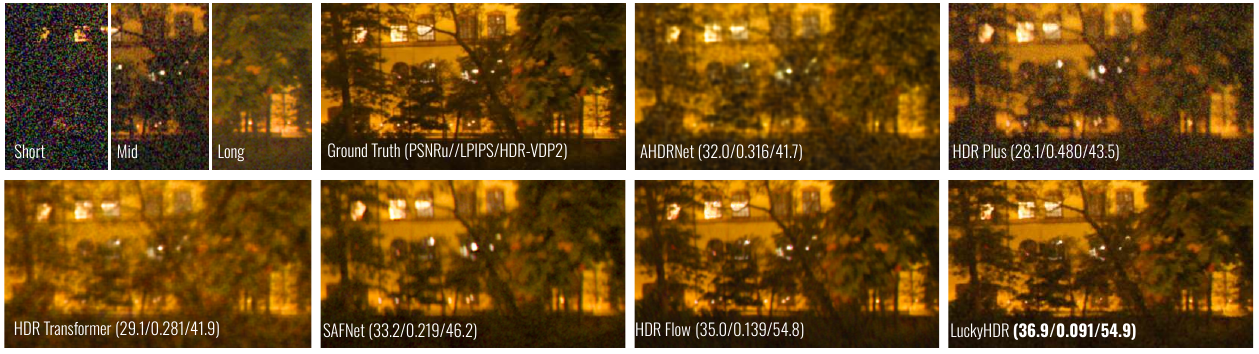}
\vspace*{-24pt}
    \caption{\rev{\textbf{Evaluation on Synthetic Data.}} We show a representative example from the SIHDR-fast synthetic test split. The left panel visualizes the three input exposures (short/mid/long). The remaining panels show the ground truth label and reconstructions from AHDRNet, HDR+, HDR-Transformer, SAFNet, HDRFlow, and LuckyHDR. While prior methods may over-smooth details or exhibit residual ghosting under exposure-tied blur and motion, LuckyHDR preserves fine structures by aligning each frame and producing the output as a convex combination of observed (shifted) input pixels. Exposure was uniformly adjusted across all images for better visualization.}
    \Description{Qualitative comparison on a synthetic HDR bracket: three input exposures, ground-truth label, and outputs from multiple baseline HDR reconstruction methods versus LuckyHDR.}
    \label{fig:synthetic}
\vspace{-3mm}
\end{figure*}

\rev{\fig{fig:synthetic} visualizes one synthetic test scene together with the three input exposures, the ground-truth label, and the reconstructions produced by every baseline we report in \tbl{tab:quan} and by LuckyHDR. We pick this scene because it contains the two regimes that separate the methods in our quantitative evaluation: bright window panes that are clipped in the mid and long exposures, and deep shadows that remain noisy in the short exposure. On this scene, HDR-Transformer and AHDRNet smear high-frequency facade detail, while HDRFlow and SAFNet leave a faint ghost around the window frame where the frames disagree under our exposure-tied blur. LuckyHDR recovers the facade structure and the window detail without either failure mode, because each output pixel is obtained by aligning and merging only observed input pixels rather than by regressing them.}
%\subsection{Qualitative and Quantitative Results}

\relatedpar{Ablation Experiments}\label{sec:ablations}
The supplement provides extensive ablation experiments that confirm key design decisions of the proposed method. Specifically, we evaluate the contribution of individual frames, the effect of resolution, base frame selection, and robustness to degradation effects. These experiments all confirm our choice of the proposed iterative align and merge strategy with the two-stage coarse-to-fine network architecture.

\relatedpar{\rev{Base Frame Selection}}
\rev{The ``base frame'' denotes the frame into which all warped and merged, so its sharpness and its highlight content upper-bound the quality of the final reconstruction. We therefore compare the two natural candidates in a 3-exposure bracket: the shortest exposure (sharpest, with the most highlight headroom but the noisiest shadows) and the middle exposure (better SNR in the shadows but more motion blur and more clipped highlights). We retrain LuckyHDR with each choice and evaluate on our SI-HDR-fast synthetic handheld split at exposure step sizes 2 and 3. \tbl{tab:frame} reports PSNR$_l$, PSNR$_\mu$, SSIM$_\mu$, HDR-VDP2, and LPIPS for all four settings. At step size 2, the shortest exposure wins on every metric by a large margin, and at step size 3 it still leads on PSNR$_l$, PSNR$_\mu$, HDR-VDP2, and LPIPS; the two variants are effectively tied on SSIM$_\mu$ (within 0.002). The gap widens on the other metrics as step size grows, because the longer exposures carry more exposure-tied blur and clip a larger fraction of the highlights that the base frame is supposed to anchor.}
\bl{\begin{table}[t]
    % Base-frame ablation (FAST_REALBLUR) with LPIPS/HDR-VDP2 (seed=46, 37 samples):
    %   hdrproj/1211/LuckyHDR/exp/baseframe_fast_realblur_seed46_lpips_hdrvdp2.json
    \caption{\textbf{Effect of Base Frame Choice.} We evaluate the performance on the synthetic dataset for different choices of base frame.
    In this experiment, we limit each scene to only three exposures and report results (PSNR$_l$, PSNR$_\mu$/SSIM$_\mu$, HDR-VDP2, LPIPS) for step sizes 2 and 3 under our exposure-tied blur simulator (shortest exposure is sharp; blur magnitude scales with exposure time).
    \textit{Shortest Base} and \textit{Middle Base} indicate that the base is the frame with shortest and middle exposure times, respectively.}
    \centering
    \footnotesize
    \vspace*{-8pt}
    \begin{tabular}{ccccccc}
      Variations & Step & PSNR$_{l}$ & PSNR$_\mu$ & SSIM$_\mu$ & HDR-VDP2 & LPIPS\\
      \midrule
        Shortest Base &2&49.5&33.8&0.823&40.7&0.226\\
        Shortest Base &3&49.6&36.0&0.913&42.1&0.106\\
        Middle Base &2&40.7&32.9&0.818&34.1&0.248\\
        Middle Base &3&36.7&34.9&0.915&36.6&0.110\\

    \end{tabular}
    \label{tab:frame}
    \vspace*{-13pt}
\end{table}}  

% \vspace{-5mm}

\begin{figure*}[t]
\centering\includegraphics[width=0.99\linewidth]{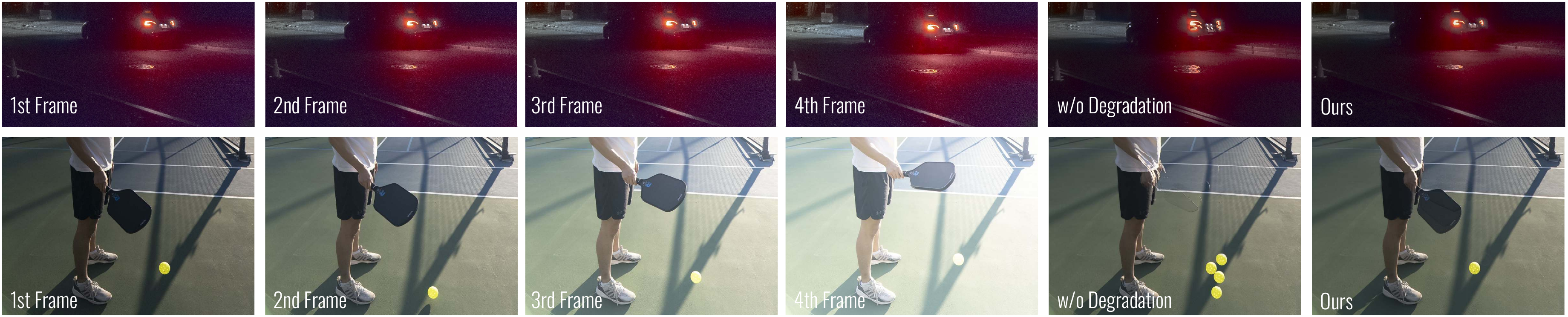}
\vspace*{-12pt}
    \caption{\textbf{Ablation on Degradation in Simulation.} We compare inference on unseen real-world data. Training LuckyHDR \emph{without} simulated degradations (e.g., foreground motion and blur) fails to handle motion at test time, producing blur (top row, car lights) and ghosting (bottom row, duplicated pickleballs). In contrast, training with our proposed augmentation handles these artifacts effectively.}
    \label{fig:degradation}
    \vspace{-8pt}
\end{figure*}

\relatedpar{\rev{Effect of Realistic Simulation in Training}} 
\rev{We retrain LuckyHDR without the simulated motion blur, translation, and noise augmentations described in \sect{sec:data}, and evaluate the two variants on the same real handheld brackets. In \fig{fig:degradation}, the ``w/o degradation'' column carries residual ghosts along the car body and a blurry halo around the moving hand, two artifacts that match what the alignment stage fails on when it has never seen exposure-tied blur and per-frame hand shake at training time. The full pipeline reconstructs the same regions cleanly, confirming that the motion and blur augmentations, rather than the architecture alone, are what closes the gap between synthetic training and real handheld captures.}

\begin{figure*}[t]
\centering\includegraphics[width=0.99\linewidth]{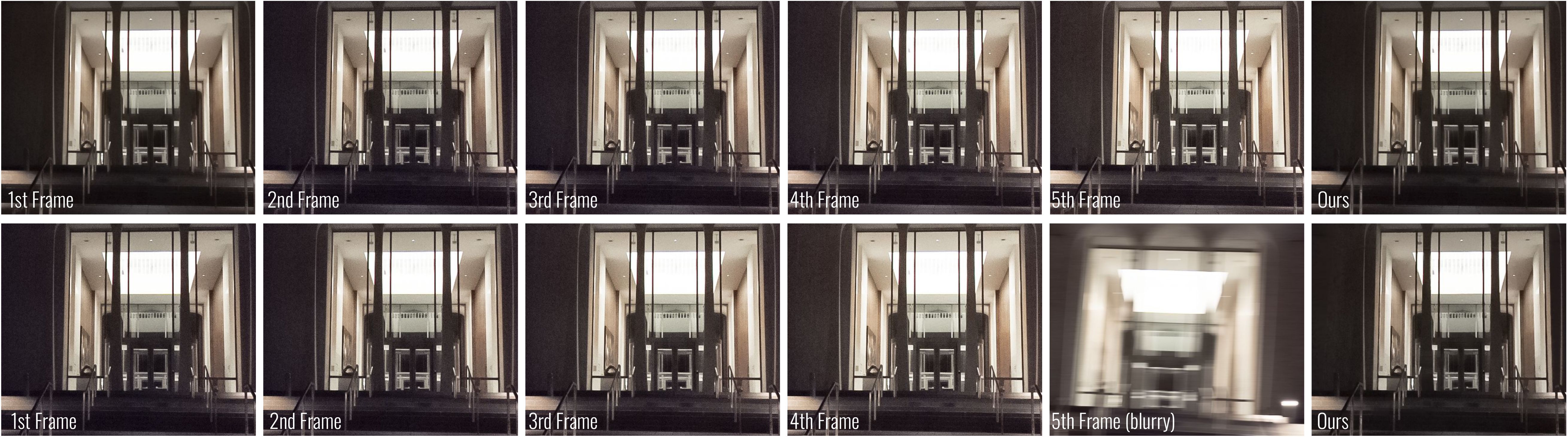}
\vspace*{-8pt}
    \caption{\textbf{Effect of Severe Hand Shake.} We assess the robustness of the method to severe blur in the 5th frame to simulate extreme hand shake. We note our approach remains robust when dealing with this edge case.}
    \label{fig:blurry}
    \vspace{-8pt}
\end{figure*}

\relatedpar{\rev{Severe Hand Shake}} 
\rev{To stress-test the merge stage, we corrupt the fifth frame of a 5-exposure bracket with a heavy direction-aligned motion blur that mimics extreme hand shake, while leaving the other four frames untouched. In \fig{fig:blurry}, the reconstruction does not inherit the blur: the predicted merge weights for the corrupted frame collapse toward zero in the affected regions, so the final pixel is drawn almost entirely from the four sharp frames, which together already cover the full dynamic range of the scene. This matches the design of the merge stage, which treats unusable frames as outliers rather than requiring them to be dropped by the user.}
%% [Stage 3] figure_gridresults_new moved inline from paper_suffix.tex.
\subsection{Experimental Handheld Captures}
\label{sec:handheld}

% Smartphone as well as big camera images
% We qualitatively evaluate
% We captured N scenes with an iPhone 12 Pro smartphone and M scenes with an SLR.
% They are intended to be ``shots worth taking'' and meant to represent situations that can happen in real life.
% Again, we do not claim to be the last word on this problem. You can always construct a harder scene where the technique falls over.

\rev{We report qualitative results on real-world handheld bracketed bursts captured with a custom iPhone app and a synthetic burst rendered with Blender. Neither dataset was used for training.}
% \rev{These captures are unseen to all evaluated methods, including the baselines we re-trained on our synthetic dataset.}
%and a Canon EOS R3 mirrorless camera. \adam{bridge? are we removing this?}
%See Supplemental Materials for additional details on our capture application.
We choose realistic but challenging scenarios where we do not expect significant scene motion.
All sequences are taken handheld without a tripod and the photographer instead strives to keep the camera as still as possible.
%
%while mirrorless captures use exposure compensation set to $(-4, -2, 0, +2, +4)$ EV.
%This is a common photography setup for both casual photographers and professionals since it is not always convenient to set up a tripod or environmental conditions negate the benefits of a tripod, for example, a windy day or a rocky boat.
Inter-frame motion will thus be a result of either slight camera shake or moving subjects. 
Smartphone captures are automatically bracketed with AE (\sect{sec:capture}).

\rev{Before model inference, RAW captures undergo minimal post-processing: each DNG is read, demosaiced, highlight-recovered, and transformed via the embedded color correction matrices into \emph{linear sRGB} space. To export model outputs as DNG files, we invert the color correction matrices and write back \emph{linear RAW} values with modified metadata, allowing subsequent rendering steps (e.g., vignetting correction) to be handled by standard RAW editing software using the embedded DNG metadata. The visualized results in the paper and supplemental material were tonemapped using the Adaptive Color mode in Adobe Camera Raw, with manual tone adjustments to enhance shadow detail.}

\rev{In \fig{fig:iphone_comparison}, we show the HEIC output from the iPhone 16 Pro Max native camera, which is post-processed internally, alongside LuckyHDR results tonemapped in Adobe Camera Raw with manual adjustments to approximate the iPhone's appearance. The two scenes are chosen to highlight contrasting approaches to HDR imaging.
In the first scene (top row), the iPhone native Camera benefits from access to higher-resolution sensor data unavailable to third-party apps, yielding sharper and cleaner text rendering, but at the cost of hallucinated detail that deforms small characters. In the second scene (bottom row), LuckyHDR more faithfully reproduces the ``Gift Shop'' sign without excessive contrast enhancement, and better preserves detail in the fluorescent lighting tubes beneath it.}

\rev{We also present qualitative comparisons against baseline models \emph{retrained} on our synthetic data. \fig{fig:mobile_comparison} presents results across five handheld captures featuring mixed light sources, moving pedestrians, and large shadow regions. AHDRNet and HDR-Transformer exhibit ghosting around moving subjects, distort shadow color, and hallucinate cross-shaped artifacts around light sources. HDRFlow and SAFNet produce overall high-quality results, but struggle around light sources and fail to align high-frequency textured regions.} Our method remains robust, successfully denoising the dark regions, effectively aligning the frames, and merging each pixel of the scene while avoiding color distortion.
% We also compare our method to HDR-Transformer~\cite{liu2022ghost} and HDRFlow~\cite{xu2024hdrflow}, the two best-performing baselines from \sect{sec:synthetic}, in Figures \ref{fig:additional_1} and \ref{fig:additional_2}.
% In these challenging scenes, ours is the only method that avoids ghosting while simultaneously reducing noise in the shadows. The proposed method recovers fine detail around the tree branches and windows while denoising dark background content and building facades.
%
% \blc{Do we still have bruno method for visualized comparison? If no then delete the following sentence.} -- AF: I moved it to Supplemental
See the  Supplemental Materials for additional real-world test results, as well as a comparison to the method of Lecouat~\etal~\shortcite{lecouat2022hdr} for daytime HDR capture. 

% See the supplement for additional results, including images suitable for HDR displays, and a selection of linear input brackets.

To further validate the robustness of the iterative align-and-merge approach we test it on a synthetic (rendered) scene involving an extreme 9-frame bracketing sequence, with simulated hand shake, straddling more than 20 camera stops, shown in \fig{fig:barbershop}. Our method is able to recover detail across the entire dynamic range (though tone mapping struggles to visualize this range), while baseline methods clip throughout the window area. %Note that our current mobile capture app is unable to capture such inputs because of the extreme range of exposures, but in principle such an app could be built and would require a patient and steady photographer.

%Nevertheless, we find that our method produces detailed results while minimizing ghosting and hallucination artifacts, owing to our fusion strategy of predicting merging weights as opposed to direct pixel prediction.
%Results are shown in Figures \ref{fig:additional_1} and \ref{fig:additional_2}. \bl{As can be seen in the figure, the comparison methods all fail to perform accurate shift, thus causing sever ghost and unsatisfactory visual experience. Our method, on the other hand, could successfully perform accurate global shift (with the help of pre-trained model) and local shift with our strategy. We further provide the inputs of our captured data, which can help to understand how our method works. }.

\vspace*{-0.8\baselineskip}
\subsection{Runtime}
We benchmark our method on desktop GPUs for offline processing, akin to photo-finishing applications, and on mobile devices. Specifically, we list the RTX A6000 inference time per output image (ITPI) at $1888 \times 1280$ in Tab.~\ref{tab:quan}. The quality/runtime tradeoff for different bracket lengths is reported in the supplement. Without substantial optimization, benchmarks on the iPhone 17 Pro Max shows that a single iteration (merging two images) takes 8.33 ms at 512×512 using the NPU while at 2048×2048, it takes 328 ms (via CPU+GPU). We additionally provide qualitative experiments on the influence of increasing the number of input frames in \rev{the Supplemental Material}.

\rev{\tbl{tab:compute} reports the inference time per image (ITPI), parameter count, and FLOPs of every evaluated method on a single RTX A6000 with 50~GB of memory, at three output resolutions: $512^2$, $2048^2$, and $4096^2$. At $512^2$, which is the resolution used for most learning-based HDR benchmarks, LuckyHDR runs in 7~ms per merge, roughly $160\times$ faster than HDR-Transformer (1160~ms), $4\times$ faster than SAFNet (28~ms), and $8\times$ faster than AHDRNet (56~ms), while using only 66K parameters -- between $17\times$ and $50\times$ fewer than the deep baselines (1.1--3.27~M). As the resolution grows, the gap widens because the transformer-based baselines scale quadratically with image area: at $2048^2$, HDR-Transformer and AFUNet already take more than 17~s and 24~s per frame respectively, against 107~ms for LuckyHDR. At $4096^2$, HDR-Transformer, AHDRNet, and AFUNet exceed the 50~GB memory budget and cannot produce an output at all, while among the deep learning baselines that still fit in memory, LuckyHDR maintains the same efficiency of HDRFlow (430~ms vs.\ 424~ms).}

%These timings are reported per iteration and reflect end-to-end inference for one merge.

% \subsection{Contribution of Individual Merged Frames}
% In addition to the quantitative results of \tbl{tab:numbers_stops}, we also visualize outputs of our method with a varying number of input frames to help analyze the iterative process of our approach in \fig{fig:number}. Given any number of inputs (1 to 5 in the example), our model feeds inputs into the network one by one and gradually improves the final result. 

%using Xcode's built-in profiler.\bl{Also we use different stops of each adjacent frames to test its generalization.

%\vspace*{-0.5\baselineskip}
\vspace{-1mm}
\section{Discussion}
\label{sec:conclusion}

\begin{figure}[t]
\vspace{-8pt}
\centering\includegraphics[width=0.99\linewidth]{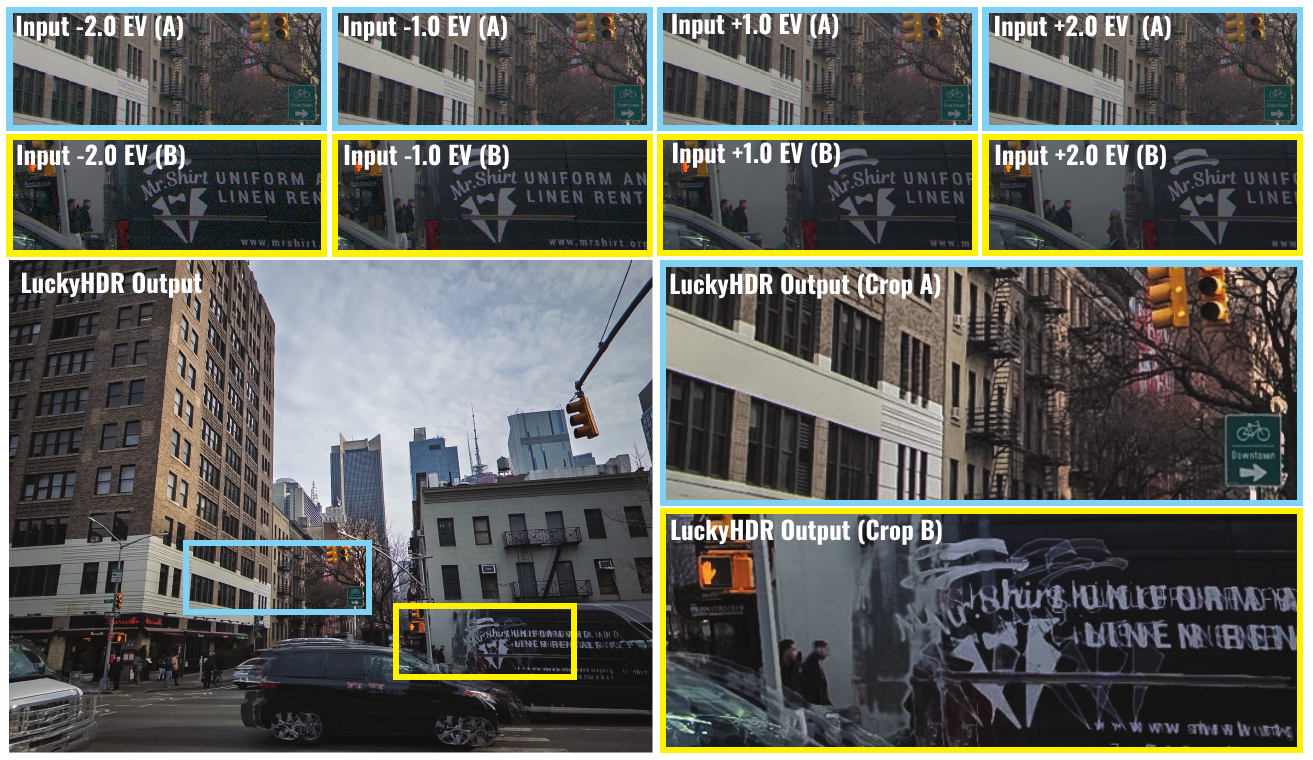}
    \vspace{-8pt}
    \caption{\rev{\textbf{Extreme Motion.} While our method handles small scene motions effectively (e.g., ~5 px of hand-shake in crop A), it fails for large local motions (e.g., ~80 px displacement from the moving vehicle in crop B), producing visible ghosting artifacts in the merged HDR output.}}
    \label{fig:failure}
\vspace{-10pt}
\end{figure}

\rev{The proposed method struggles with large local motion in low-texture regions (e.g., fast-moving pedestrians or vehicles; see Figure~\ref{fig:failure}), which is difficult to align with lightweight networks and to robustly reject. We intentionally exclude these scenarios from our scope, as we assume a mode of intentional photography analogous to a camera's ``night mode,'' in which the user is expected to hold the camera still to capture a relatively static scene. A practical extension of our method would be to preprocess the burst with feature-based homography alignment, reducing large inter-frame displacements to small residual misalignments (around 8 px) that our alignment network can handle reliably. The merge network could then more effectively identify and suppress the remaining misaligned regions.}

\rev{Several smartphone ISP effects are not modeled in either the capture model or synthetic burst generation. For example, exposure-dependent PSF changes induce a mild per-frame blur, and veiling glare around light sources leaves an additive light contribution even after exposure normalization. Though these effects tend to be small, explicitly modeling them would be a natural direction for future work. Additionally, our method assumes a fixed ISO across the bracketed burst, while jointly varying ISO and exposure within a burst schedule could further extend captured dynamic range.}

\section{Conclusion}
\label{sec:conclusion}

We propose a lightweight alignment and merging method for fusing bracketed images spanning a large dynamic range. It operates at interactive rates for full-resolution smartphone images, allowing it to be run as an alternative photographic mode on mobile devices. Our method predicts spatially varying weight maps that are used to fuse the frames from the burst. By design, pixel values in the final image can only be constructed as a weighted combination of pixels from the input stack which makes it unlikely to hallucinate image content. %But unlike earlier works on HDR weight prediction such as HDR+~\cite{hasinoff2016burst} our method leverages advances in deep learning, specifically transformer based attention scoring, to predict more sophisticated weight maps that can adapt better to diverse image content. 
This formulation allows us to train lightweight neural networks for the alignment and merging tasks that are iteratively run on the burst from short to long exposure.
We validate the method in simulation and on unseen real-world data and confirm that the approach generalizes. \rev{Tackling large dynamic motion in future work may rely on self-supervision of the burst measurements to improve alignment accuracy in the complex real-world scenarios.}

%\newpage
\bibliographystyle{ACM-Reference-Format}
\bibliography{references}
% \newpage
% \input{figs_and_tabs/figure_gridresults_new}
% \clearpage
% \input{figs_and_tabs/figure_synthetic}

% \input{figs_and_tabs/figure_iphone_new}
% \clearpage

% \newpage
% \bibliographystyle{ACM-Reference-Format}
% \bibliography{references}

\end{document}